\documentclass{article}

     \PassOptionsToPackage{numbers, compress}{natbib}
     \usepackage[preprint]{neurips_2019}

\usepackage[utf8]{inputenc} 
\usepackage[T1]{fontenc}    
\usepackage{hyperref}       
\usepackage{url}            
\usepackage{booktabs}       
\usepackage{multirow}
\usepackage{amsfonts, amsmath}       
\usepackage{nicefrac}       
\usepackage{microtype}      
\usepackage{graphicx}
\usepackage{bm}
\usepackage{subfig}
\usepackage{tikz} 

\title{Stochastic Bottleneck: Rateless Auto-Encoder for Flexible Dimensionality Reduction}

\author{%
  Toshiaki Koike-Akino and Ye Wang\\
  Mitsubishi Electric Research Laboratories (MERL)\\
  201 Broadway, Cambridge, MA 02139\\
  \texttt{koike@merl.com, yewang@merl.com} \\
}

\begin{document}

\maketitle

\begin{abstract}
We propose a new concept of rateless auto-encoders (RL-AEs) that enable a flexible latent dimensionality,
which can be seamlessly adjusted for varying distortion and dimensionality requirements.
In the proposed RL-AEs, instead of a deterministic bottleneck architecture,
we use an over-complete representation that is stochastically regularized with weighted dropouts, in a manner analogous to sparse AE (SAE).
Unlike SAEs, our RL-AEs employ monotonically increasing dropout rates
across the latent representation nodes such that the latent variables become sorted by importance
like in principal component analysis (PCA). This is motivated by the rateless property
of conventional PCA, where the least important
principal components can be discarded to realize variable rate dimensionality
reduction that gracefully degrades the distortion.
In contrast, since the latent variables of conventional AEs are equally important
for data reconstruction, they cannot be simply discarded to further reduce the dimensionality after the AE model is
trained.
Our proposed stochastic bottleneck framework enables seamless rate adaptation with high reconstruction performance,
without requiring predetermined latent dimensionality at training.
We experimentally demonstrate that the proposed RL-AEs can
achieve variable dimensionality reduction while achieving low distortion compared to conventional AEs.
\end{abstract}

\section{Introduction}

In many real-world applications, the raw data measurements (e.g., audio/speech, images, video, biological signals)
often have very high dimensionality.
Adequately handling high-dimensionality often requires the application of
dimensionality reduction techniques~\cite{Maaten:2009} that transform the
original data into meaningful feature representations
of reduced dimensionality. Such feature representations should reduce the
dimensionality to the minimum number required to capture the salient
properties of the data. Dimensionality reduction is vital
in many machine learning applications, since one needs to
mitigate the so-called ``curse of dimensionality''~\cite{Jimnez:1997}.
In the past few decades, latent representation learning
based on auto-encoders (AEs)~\cite{Hinton:2006, Schloz:2008,
Kramer:1991, DeMers:1993, Ng:2011, Vincent:2010, Doersch:2016,
Sonderby:2016} has been widely used for dimensionality reduction,
since this nonlinear technique has shown superior real-world performance compared to
classical linear counterparts, such as principal component analysis
(PCA).

One of the challenges in dimensionality reduction is to determine the optimal
latent dimensionality that can sufficiently capture the data
features required for particular applications.
Although some regularization techniques, such as sparse AE (SAE)~\cite{Ng:2011}
and rate-distortion AE~\cite{Giraldo:2013}, may be useful to self-adjust the effective dimensionality,
there are no existing methods that provide a \textit{rateless} property~\cite{MacKay:2005}
that allows for seamlessly adjustment of the latent dimensionality depending on varying distortion requirements for different downstream applications, without modification of the trained AE model.
However, realizing a rateless AE is not straightforward, since traditional AEs typically learn nonlinear manifolds where the latent variables are equally important, unlike the linear manifold models used for PCA.

In this paper, we introduce a novel AE framework which can universally achieve flexible dimensionality reduction while achieving high performance.
Motivated by the fact that the traditional PCA is readily adaptable to any dimension by just appending or dropping sorted principal components,
we propose a stochastic bottleneck architecture to associate upper latent variables with higher-principal nonlinear features so that the user can freely discard the least-principal latent variables if desired.
Our contributions are summarized below:
\begin{itemize}
 \item We introduce a new concept of rateless AEs designed for flexible dimensionality reduction.
 \item A stochastic bottleneck framework is proposed to prioritize the latent space non-uniformly.
 \item An extended regularization technique called TailDrop is considered to realize rateless AEs.
 \item We discuss dropout distribution optimization under the principle of multi-objective learning.
 \item We demonstrate that the proposed AEs achieve excellent distortion performance over the variable range of dimensionality in the standard MNIST and CIFAR-10 image datasets.
 \item We evaluate AE models trained for a perceptual distortion measure based on structural similarity (SSIM)~\cite{Wang:2004} as well as the traditional mean-square error (MSE) metric.
\end{itemize}

\section{Rateless auto-encoder (RL-AE)}

\subsection{Dimensionality reduction}

\begin{figure}[t]
 \centering
 \includegraphics[width=\linewidth]{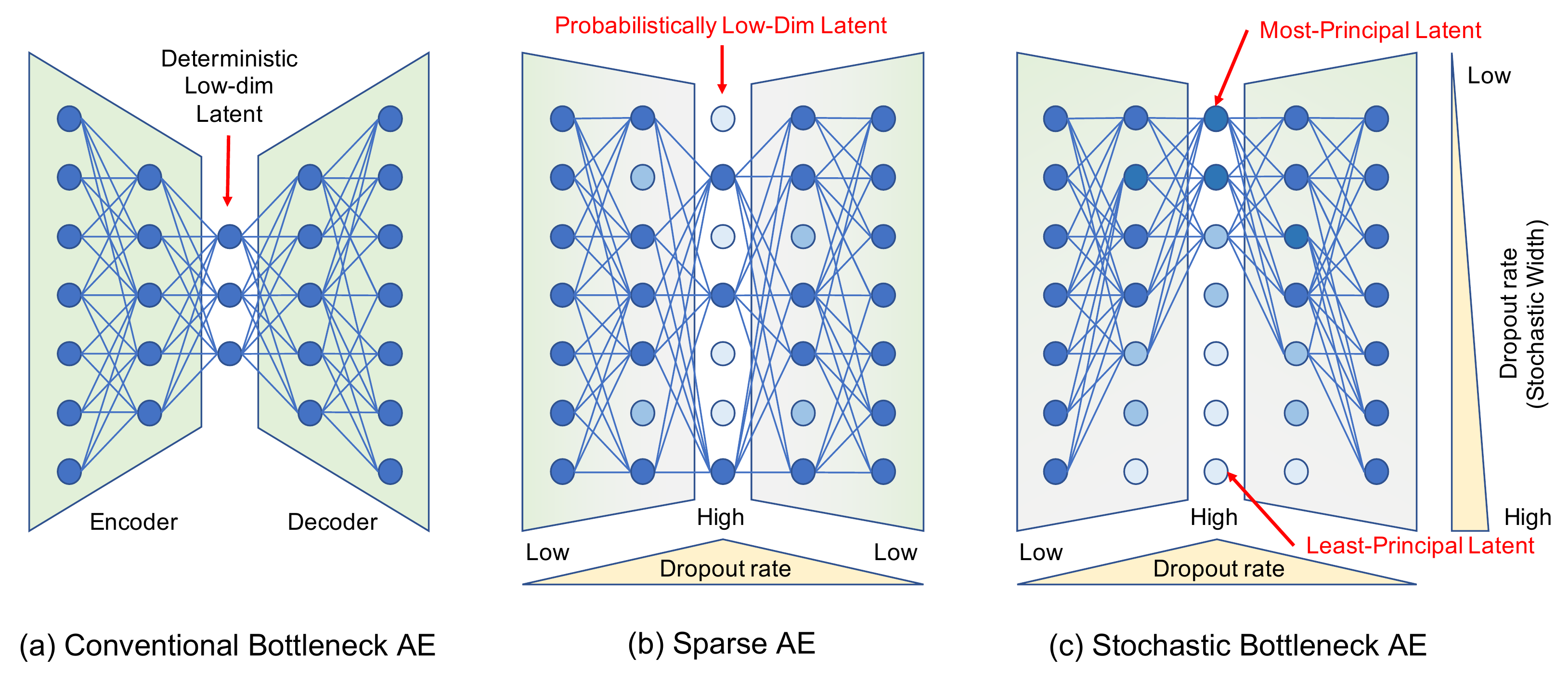}
 \caption{Auto-encoder architectures: (a) conventional bottleneck network, (b) sparse AE regularized by dropout, having probabilistically lower-dimension representation, (c) stochastic bottleneck, having two-dimensional regularization with non-identical dropout rates in both depth and width directions to realize rateless property by ordered-principal latent variables.}
 \label{fig:model}
\end{figure}

Due to the curse of dimensionality, representation learning to reduce the dimensionality is often of great importance to handle high-dimensional datasets in machine learning.
To date, there have existed many algorithms for dimensionality reduction~\cite{Maaten:2009}, e.g., PCA, kernel PCA, Isomap, maximum variance unfolding, diffusion maps, locally linear embedding, Laplacian eigenmaps, local tangent space analysis, Sammon mapping, locally linear coordination and manifold charting along with AE.
Among all, AE~\cite{Hinton:2006, Schloz:2008, Kramer:1991, DeMers:1993, Ng:2011, Vincent:2010, Doersch:2016, Sonderby:2016} has shown its high potential to learn lower-dimensional latent variables required in the nonlinear manifold underlying the datasets.
AE is an artificial neural network having a bottleneck architecture as illustrated in Fig.~\ref{fig:model}(a), where $N$-dimensional data is transformed to $M$-dimensional latent representation (for $M \leq N$) via an encoder network.
The latent variables should contain sufficient feature capable of reconstructing the original data through a decoder network.

From the original data $\mathbf{x} \in \mathbb{R}^{N}$,
the corresponding latent representation $\mathbf{z} \in \mathbb{R}^{M}$, with a reduced dimensionality $M$ is generated by the encoder network as $\mathbf{z} = f_\theta(\mathbf{x})$, where $\theta$ denotes the encoder network parameters.
The latent variables should adequately capture the statistical geometry of the data manifold, such that 
the decoder network can reconstruct the data as $\mathbf{x}' = g_\phi(\mathbf{z})$, where $\phi$ denotes the decoder network parameters and $\mathbf{x}' \in \mathbb{R}^{N}$.
The encoder and decoder pair $(f_\theta, g_\phi)$ are jointly trained to minimize the reconstruction loss (i.e., distortion), as given by:
\begin{align}
 \mathop{\min}_{\theta, \phi}\,
 \mathop{\mathbb{E}}_{\mathbf{x} \sim \Pr(\mathbf{x})}
 \Big[
 \mathcal{L}\big(\mathbf{x}, g_\phi ( f_\theta ( \mathbf{x} ) )\big)
 \Big],
 \label{eq:obj}
\end{align}
where the loss function $\mathcal{L}(\mathbf{x}, \mathbf{x}')$ is chosen to quantify the distortion (e.g., MSE) between $\mathbf{x}$ and $\mathbf{x}'$.

\subsection{Motivation: rateless property}

By analogy, AEs are also known as nonlinear PCA (NLPCA)~\cite{Schloz:2008, Kramer:1991, DeMers:1993}.
If we consider a simplified case where there is no nonlinear activation in the AE model, then the encoder and decoder functions will reduce to simple affine transformations.
Specifically, the encoder becomes $f_\theta(\mathbf{x}) = \mathbf{W} \mathbf{x} + \mathbf{b}$ where trainable parameters $\theta$ are the linear weight $\mathbf{W} \in \mathbb{R}^{M\times N}$ and the bias $\mathbf{b} \in \mathbb{R}^{M}$.
Likewise, the decoder becomes $g_\phi(\mathbf{z}) = \mathbf{W}' \mathbf{z} + \mathbf{b}'$ with $\phi = \{\mathbf{W}', \mathbf{b}'\} \in \{ \mathbb{R}^{N\times M}, \mathbb{R}^{N} \}$.
If the distortion measure is MSE, then the optimal linear AE coincides with the classical PCA when the data follows the multivariate Gaussian distribution according to the Karhunen--Lo\`{e}ve theorem.

To illustrate, assume Gaussian data $\mathbf{x} \sim \mathcal{N}(\mathbf{m}, \mathbf{C})$ with mean $\mathbf{m} \in \mathbb{R}^{N}$ and covariance $\mathbf{C} \in \mathbb{R}^{N\times N}$, which has the eigen-decomposition: $\mathbf{C} = \bm{\Phi} \bm{\Lambda} \bm{\Phi}^\mathrm{T}$, where $\bm{\Phi} \in \mathbb{R}^{N\times N}$ is the unitary eigenvectors matrix and $\bm{\Lambda} = \mathsf{diag}[\lambda_1, \lambda_2, \ldots, \lambda_N] \in \mathbb{R}^{N\times N}$ is the diagonal matrix of ordered eigenvalues $\lambda_1 \geq \lambda_2 \geq \cdots \geq \lambda_N \geq 0$.
For PCA, the encoder uses $M$ principal eigenvectors $\bm{\Phi} \mathbf{I}_{N,M}$ to project the data onto an $M$-dimensional latent subspace with $\mathbf{W} = \mathbf{I}_{M,N} \bm{\Phi}^\mathrm{T}$ and $\mathbf{b} = -\mathbf{W}\mathbf{m}$, where $\mathbf{I}_{M,N} \in \mathbb{R}^{M\times N}$ denotes the incomplete identity matrix with diagonal elements equal to one and zero elsewhere.
The decoder uses the transposed projection with $\mathbf{W}' = \bm{\Phi} \mathbf{I}_{N,M}$ and $\mathbf{b}' = \mathbf{m}$.
The MSE distortion is given by
\begin{align}
 \bar{\mathcal{L}}_M &=
 \mathbb{E}_\mathbf{x}
 \Big[
 \big\| \mathbf{W'} (\mathbf{W} \mathbf{x} + \mathbf{b}) + \mathbf{b}' - \mathbf{x} \big\|^2
 \Big]
 =
 \sum_{n=M+1}^{N} \lambda_n.
 \label{eq:eigen}
\end{align}
Since the eigenvalues are sorted, the distortion gracefully degrades as principal components are removed in the corresponding order.
Of course, the MSE would be considerably worse if an improper ordering (e.g., reversed) is used.

One of the benefits of classical PCA is its graceful \textit{rateless} property due to the ordering of principal components.
Similar to rateless channel coding such as fountain codes~\cite{MacKay:2005}, PCA does not require a pre-determined compression ratio $M/N$ for dimensionality reduction (instead it can be calculated with $M=N$), and the latent dimensionality can be later freely adjusted depending on the downstream application.
More specifically, the PCA encoder and decoder learned for a dimensionality of $M$ can be universally used for any lower-dimensional PCA of latent size $L \leq M$ without any modification of the PCA model but simply dropping the least-principal $D$ components ($D=M-L$) in $\mathbf{z} = [z_1, z_2, \ldots, z_M]^\mathrm{T}$, i.e., nullifying the tail variables as $z_m = 0$ for all $m \in \{L+1, \ldots, M\}$.

The rateless property is greatly beneficial in practical applications since the optimal latent dimensionality is often not known beforehand. 
Instead of training multiple encoder and decoder pairs for different compression rates, one common PCA model can cover all rates $L/N$ for $1\leq L \leq M$ by simply dropping trailing components, while still attaining good performance as given by $\bar{\mathcal{L}}_L$.
For example, a medical institute could release a massively high-dimensional magnetic resonance imaging (MRI) dataset alongside a trained PCA model with a reduced-dimensionality of $M$ targeted for a specific diagnostic application. 
However, for under various other applications (e.g., different analysis or diagnostic contexts), an even further reduced dimensionality may suffice and/or improve learning performance for the ultimate task.
Even for end-users that require fewer latent variables in various applications, the excellent rate-distortion tradeoff (under Gaussian data assumptions) is still achieved, without updating the PCA model, by simply discarding the least-principal components.

Nevertheless, traditional PCA often underperforms in comparison to nonlinear dimensionality reduction techniques on real-world datasets.
Exploiting nonlinear activation functions such as rectified linear unit (ReLU), AEs can better learn inherent nonlinearities of the latent representations underlying the data.
However, existing AEs do not readily achieve the rateless property, because the latent variables are generally learned to be equally important.
Hence, multiple AEs would need to be trained and deployed for different target dimensionalities.
This drawback still holds for the progressive dimensionality reduction approaches employed by stacked AEs~\cite{Hinton:2006} and hierarchical AEs~\cite{Schloz:2008}, those of which require multiple training and re-tuning for different dimensionality.
In this paper, we propose a simple and effective technique of employing a stochastic bottleneck to realize rateless AEs that are adaptable to any compression rates.

\subsection{StochasticWidth bottleneck}

Several variants of AE have been proposed, e.g., sparse AE (SAE)~\cite{Ng:2011}, variational AE (VAE)~\cite{Vincent:2010, Doersch:2016, Sonderby:2016}, rate-distortion AE~\cite{Giraldo:2013}, and compressive AE~\cite{Theis:2017}.
We introduce a new AE family which has no fixed bottleneck architecture to realize the rateless property for seamless dimensionality reduction.
Our method can be viewed as an extended version of SAE, similar in its over-complete architecture, but also employing a varying dropout distribution across the width of the network.
This aspect of our approach is key for achieving good reconstruction performance while allowing a flexibly varying compression rate for the dimensionality reduction.

Unlike a conventional AE with a deterministic bottleneck architecture, as shown in Fig.~\ref{fig:model}(a), the SAE employs a probabilistic bottleneck with an effective dimensionality that is stochastically reduced by dropout, as depicted in Fig.~\ref{fig:model}(b).
For example, the SAE encoder generates $M$-dimensional variables $\mathbf{z}$ which are randomly dropped out at a probability of $p$, resulting in an effective latent dimensionality of $\bar{L}=(1-p)M$.
Although the SAE has better adaptability than deterministic AE to further dimensionality reduction by dropping latent variables, the latent variables are still trained to be equally important for reconstruction of the data, and thus it is limited in achieving flexible ratelessness.

Our AE employs a stochastic bottleneck that imposes a specific dropout rate distribution that varies across both the width and depth of the network, as shown in Fig.~\ref{fig:model}(c).
In particular, our StochasticWidth technique employs a monotonically increasing dropout rate from the head (upper) latent variable nodes to the tail (lower) nodes in order to encourage the latent variables to be ordered by importance, in a manner analogous to PCA.
By concentrating more important features in the head nodes, we hope to enable adequate data reconstruction even when some of the least important dimensions (analogous to least-principal components) are later discarded.

This non-uniform dropout rate may also offer another benefit for gradient optimization.
For existing AEs, the distortion is invariant against node permutations with permuted weights and bias in neural networks,
which implies that there are a large number of global solutions minimizing the loss function.
A plurality of solutions may distract the stochastic gradient, while non-uniform dropout rates can give a particular priority at every node that prevents permutation ambiguity.

\subsection{TailDrop regularization}

Dropout~\cite{Hinton:2012, Srivastava:2014} has been widely used to regularize over-parameterized deep neural networks.
The role of dropout is to improve generalization performance by preventing activations from becoming strongly correlated, which in turn leads to over-training.
In the standard dropout implementation, network activations are discarded (by zeroing the activation for that neuron node) during training (and testing for some cases) with independent probability $p$.
A recent theory~\cite{Gal:2016} provides a viable interpretation of dropout as a Bayesian inference approximation.

There are many related regularization methods proposed in literature; e.g., DropConnect~\cite{Wan:2013}, DropBlock~\cite{Wu:2018}, StochasticDepth~\cite{Huang:2016}, DropPath~\cite{Larsson:2016}, ShakeDrop~\cite{Yamada:2018}, SpatialDrop~\cite{Tompson:2015}, ZoneOut~\cite{Krueger:2016}, Shake-Shake regularization~\cite{Gastaldi:2017}, and data-driven drop~\cite{Huang:2017}.
In order to facilitate the rateless property for stochastic bottleneck AE architectures, we introduce an additional regularization mechanism referred to as TailDrop, as one realization of StochasticWidth.

\begin{figure}[t]
 \centering
 \includegraphics[width=\linewidth]{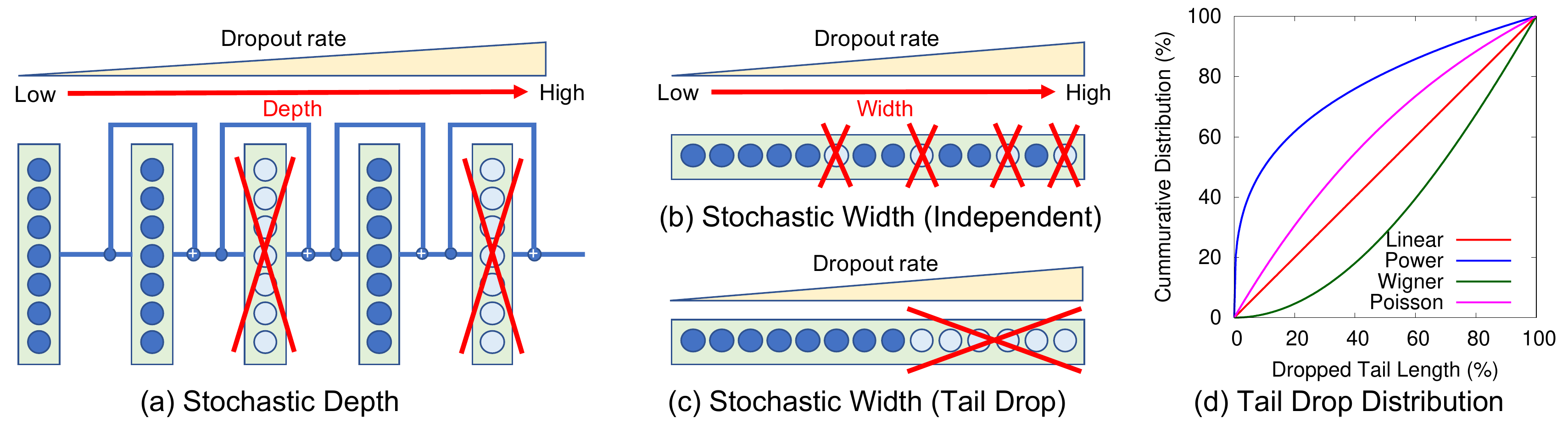}
 \caption{Non-uniform dropout regularization: (a) StochasticDepth~\cite{Huang:2016} to control depth by prioritizing shallower layers, (b) StochasticWidth to control width by prioritizing head neurons with independent and increasing dropout, (c) StochasticWidth to nullify consecutive burst neurons by TailDrop, (d) example of tail drop distributions.}
 \label{fig:drop}
\end{figure}

The stochastic bottleneck uses non-uniform dropout to adjust the importance of each neuron as explained in Fig.~\ref{fig:model}(c).
This regularization technique is related to StochasticDepth~\cite{Huang:2016} used in deep residual networks.
As illustrated in Fig.~\ref{fig:drop}(a), StochasticDepth drops out entire layers at a higher chance when dropping deeper layers so that an effective network depth is constrained and shallower layers are dominantly trained.
Analogously, non-uniform dropouts are carried out across the width direction for StochasticWidth as shown in Fig.~\ref{fig:drop}(b), where independent dropouts at increasing rates are used for each neuron.
The monotonically increasing dropout rates can be also realized by dropping consecutive nodes at the tail as shown in Fig.~\ref{fig:drop}(c), which we call TailDrop.
For TailDrop, the desired dropout rates can be achieved by adjusting the probability distribution of the tail drop length as depicted in Fig.~\ref{fig:drop}(d).
Considering the scenarios that the user would discard the least-principal latent variables to adjust dimensionality later, we focus on the use of this TailDrop regularization for rateless AE in this paper.

\subsection{Multi-objective learning}

Finding an appropriate dropout probability distribution is a key consideration in the design of high-performance rateless AEs.
We now give offer insights on how to do so, however a rigorous theoretical development remains an open problem for future study.
The objective function in~\eqref{eq:obj} should be re-formulated to realize the rateless property.
Our ultimate goal is to find AE model parameters $\theta$ and $\phi$ that simultaneously minimize distortion across multiple rates.
Specifically, this problem is an $M$-ary multi-objective optimization as follows:
\begin{align}
 \mathop{\min}_{\theta, \phi}\,
 \Big[
 \bar{\mathcal{L}}(\theta, \phi ; 1),
 \bar{\mathcal{L}}(\theta, \phi ; 2),
 \ldots,
 \bar{\mathcal{L}}(\theta, \phi ; M)
 \Big],
 \label{eq:multi}
\end{align}
where $\bar{\mathcal{L}}(\theta, \phi ; L)$ denotes the expected distortion for the candidate AE model parameterized by $\theta$ and $\phi$, given that the $M$-dimensional latent variables $\mathbf{z}$ are further reduced to $L$-dimensional variables by dropping the last $D = M -L$ variables.
In this multi-objective problem, optimizing an AE to minimize one component of the loss objective, i.e., $\bar{\mathcal{L}}(\theta, \phi ; L)$ for a particular dimensionality $L$, generally does not yield the optimal model for other dimensionalities $L' \neq L$.
Hence, a rateless AE model must account for the best balance across multiple dimensionalities in order to approach the Pareto-front solutions.

One commonly used na\"{i}ve method in multi-objective optimization is a weighted sum optimization to reduce the problem to a single objective function as follows:
\begin{align}
 \mathop{\min}_{\theta, \phi}\,\sum_{L=1}^{M} \omega_L \bar{\mathcal{L}}(\theta, \phi; L),
 \label{eq:weight}
\end{align}
with some weights $\omega_L \geq 0$.
One may choose the weights to scale the distortion to a similar amplitude as $\omega_L \simeq 1 / \bar{\mathcal{L}}^\star (\theta, \phi; L)$ for positive distortions where $ \bar{\mathcal{L}}^\star (\theta, \phi; L)$ denotes the ground solution.
As the expected distortion may depend on the eigenvalues as shown in (\ref{eq:eigen}), understanding the nonlinear eigenspectrum can facilitate in optimizing the weight distributions.
The stochastic TailDrop regularization at training phase can be interpreted as a weight $\omega_L$ since
the conventional single-objective optimization in~\eqref{eq:obj} will effectively become the weighted sum optimization in~\eqref{eq:weight}.
Accordingly, the weights will be the survivor length probability, i.e., the TailDrop distribution is $\Pr(D=M-L) = \omega_{L}$.

Besides the weighted sum approach, there are several improved methods in multi-objective optimizations such as the weighted metric method.
We leave such an optimization framework for future work.
In this paper, we consider parametric eigenspectrum assumptions for simplicity.
Under a model-based approach of nonlinear eigenspectrum assumptions, we evaluated several parametric distributions for TailDrop probability, e.g., Poisson, Laplacian, exponential, sigmoid, Lorentzian, polynomial, and Wigner distributions, some of which are depicted in Fig.~\ref{fig:drop}(d).
Through a preliminary experiment, it was found that the power cumulative distribution function $\Pr(D < \tau M) = \tau^\beta$ for an order of $\beta\simeq 1$ ($\tau$ denotes a compression rate) performed well for most cases.
Accordingly, we focus on the use of the power distribution for TailDrop in the experiments below.


\section{Experiments}

To demonstrate the principle-of-concept benefits of our rateless AEs, we use standard image datasets of MNIST and CIFAR-10~\cite{Krizhevsky:2009}.
MNIST contains handwritten $10$-class gray-scale images of size $28$-by-$28$, and thus the raw data dimensionality is $N=28^2=784$.
The dataset has $60{,}000$ and $10{,}000$ images for training and testing, respectively.
CIFAR-10 is a dataset of $32$-by-$32$ color images, representing $10$ classes of natural scene objects.
The raw data dimensionality is thus $N=32^2\times 3=3072$.
The training set and test set contain $50{,}000$ and $10{,}000$ images, respectively.

The AE models were implemented using the Chainer framework~\cite{Tokui:2015}.
For simplicity, we use fully-connected three-layer perceptron with ReLU activation functions for both encoder and decoder networks.
Note that the concept of StochasticWidth regularization to realize ordered-principal feature can be applied to recurrent and convolutional networks in a straightforward manner.
The number of nodes in the hidden layers is $1024$ for MNIST and $2048$ for CIFAR-10.
For conventional SAE, we used $90\%$ sparsity as a baseline to evaluate the robustness of flexible latent dimensionality.
Model training was performed using the adaptive momentum (Adam) stochastic gradient descent method~\cite{Kingma:2014} with a learning rate of $0.001$, and a mini-batch size of $100$.
The maximum number of epochs is $500$ while early stopping with a patience of $20$ was applied.

\subsection{MSE measure}

\begin{figure}[t]
 \centering
 \subfloat[MNIST]{\includegraphics[width=0.50\linewidth]{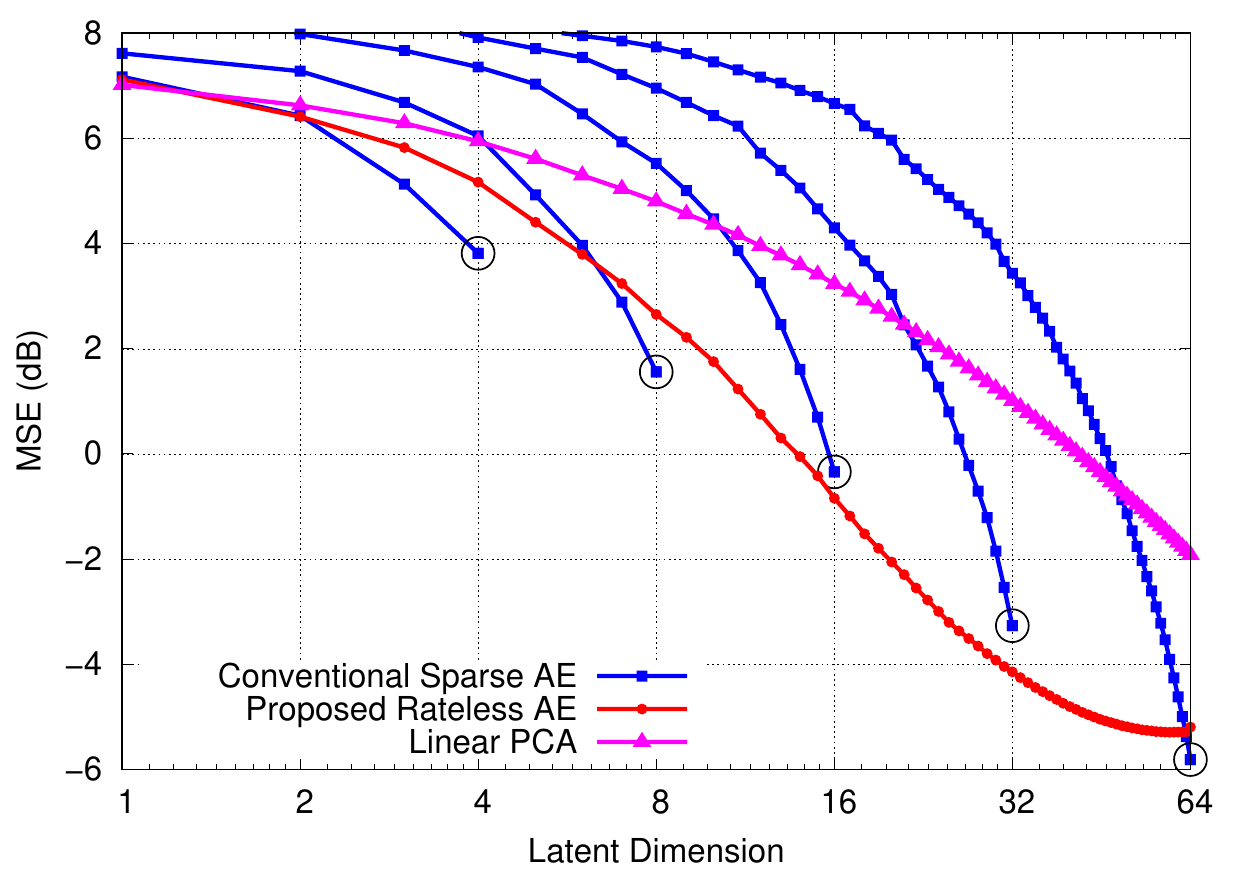}}
 \subfloat[CIFAR-10]{\includegraphics[width=0.50\linewidth]{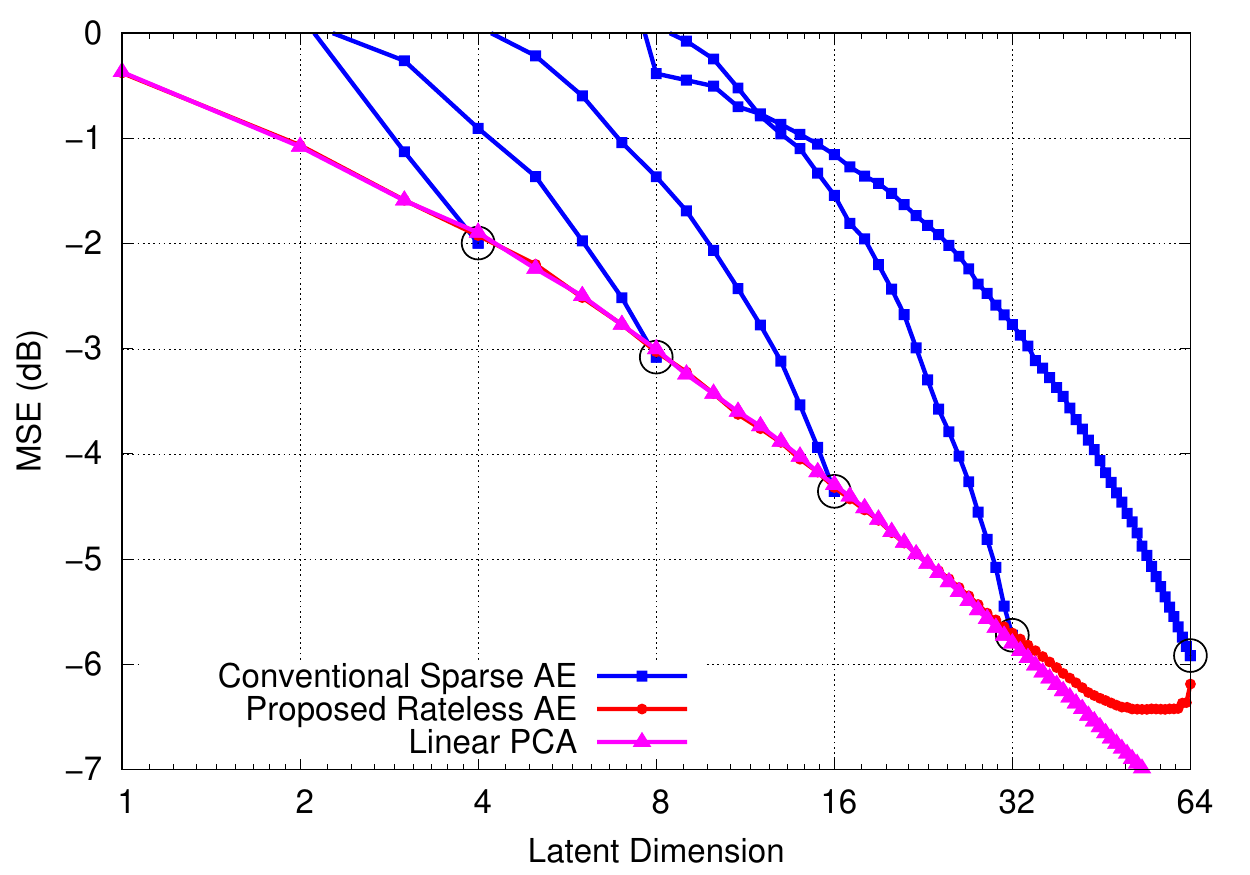}}
 \caption{MSE performance of SAE and RL-AE as a function of survivor latent dimensionality $L$.}
 \label{fig:mse}
\end{figure}

Figs.~\ref{fig:mse}(a) and (b) show the MSE performance of the conventional SAE and proposed RL-AE for MNIST and CIFAR-10 datasets, respectively.
For conventional SAE, multiple AE models are trained at the intended latent dimensionality of $M=2^m$ for $m \in \{2, 3, 4, 5, 6\}$.
The rateless AE is optimized at the dimensionality of $M=64$ using TailDrop with a power distribution (with $\beta=0.67$ for MNIST, and $\beta=2.1$ for CIFAR-10).
The parameter $\beta$ was chosen from a finite set between $0.5$ and $2.1$ to achieve a good rate-distortion tradeoff.
The latent dimensionality $L$ used for image reconstruction is varied during testing evaluation by deterministically dropping tail variables.

As shown in Fig.~\ref{fig:mse}(a), the conventional AE does not adapt well to variable dimensionality, with the MSE performance drastically degrading when the testing dimensionality $L$ is reduced from the intended dimensionality $M$.
For the SAE model trained for $M=64$, dropping $50\%$ of the latent variables to yield a reduced dimensionality of $L=32$, the MSE degrades to $3.5$~dB from the $-6$~dB obtained at $L=64$, which is significantly worse than an SAE model trained for $M=32$ that obtains an MSE of $-3.5$~dB. 
This shows that the existing SAEs cannot be universally reused for flexible dimensionality reduction, and hence adaptive switching between multiple trained SAE models would be required depending on the desired dimensionality.
However, our proposed RL-AE, which is trained once for dimensionality $M=64$, flexibly operates over the wide range of further reduced dimensionalities $L \leq 64$, while achieving low MSE distortion close to the ideal MSEs obtained by SAE models trained for the specific dimensionality $L$.

Similar observations can be made in the results for the CIFAR-10 dataset, as shown in Fig.~\ref{fig:mse}(b).
It confirms that high performance can be achieved by a single AE model for different compression rates by using the stochastic bottleneck regularization.
This benefit comes from non-uniform dropout rates across neurons to concentrate the most-principal feature in upper nodes.
Conventional uniform-rate dropout, as used in existing SAEs, still requires the target dimensionality to be known during training. 

It should be noted that the linear PCA dimensionality reduction performs surprisingly well, competitive to the proposed nonlinear AE for CIFAR-10 datasets in Fig.~\ref{fig:mse}(b).
Because MNIST images are nearly binary bitmaps whose statistics are far from the Gaussian distribution, PCA did not work well as shown in Fig.~\ref{fig:mse}(a).
However, most natural images such as CIFAR-10 are often well-modeled by the Gauss--Markov process. This may be the primary reason why PCA works sufficiently well in particular for the MSE metric.
Although it was unexpected that the nonlinear AEs could not improve the MSE performance over the linear PCA for CIFAR-10 datasets,
the MSE curve of our AE perfectly agreed that of PCA for $L\leq32$, which implies that our stochastic bottleneck approach could learn the ordered-principal components as intended.

\subsection{SSIM measure}

Here, we verify that the advantage of our rateless AEs extends beyond the MSE distortion criterion.
Since the classical MSE metric is known to be inconsistent with perceptual image quality, the structural similarity (SSIM) index~\cite{Wang:2004} has been recently used as an alternative measure of perceptual distortion.
The SSIM index ranges from $-1$ to $1$, indicating perceptual similarity between the original and distorted images, from the worst to best quality, respectively.
We use a negative SSIM index as a new loss function to fine-tune the AE models, which were pre-trained for the MSE metric, so as to improve the perceptual image quality.

\begin{figure}[t]
 \centering
 \subfloat[MNIST]{\includegraphics[width=0.49\linewidth]{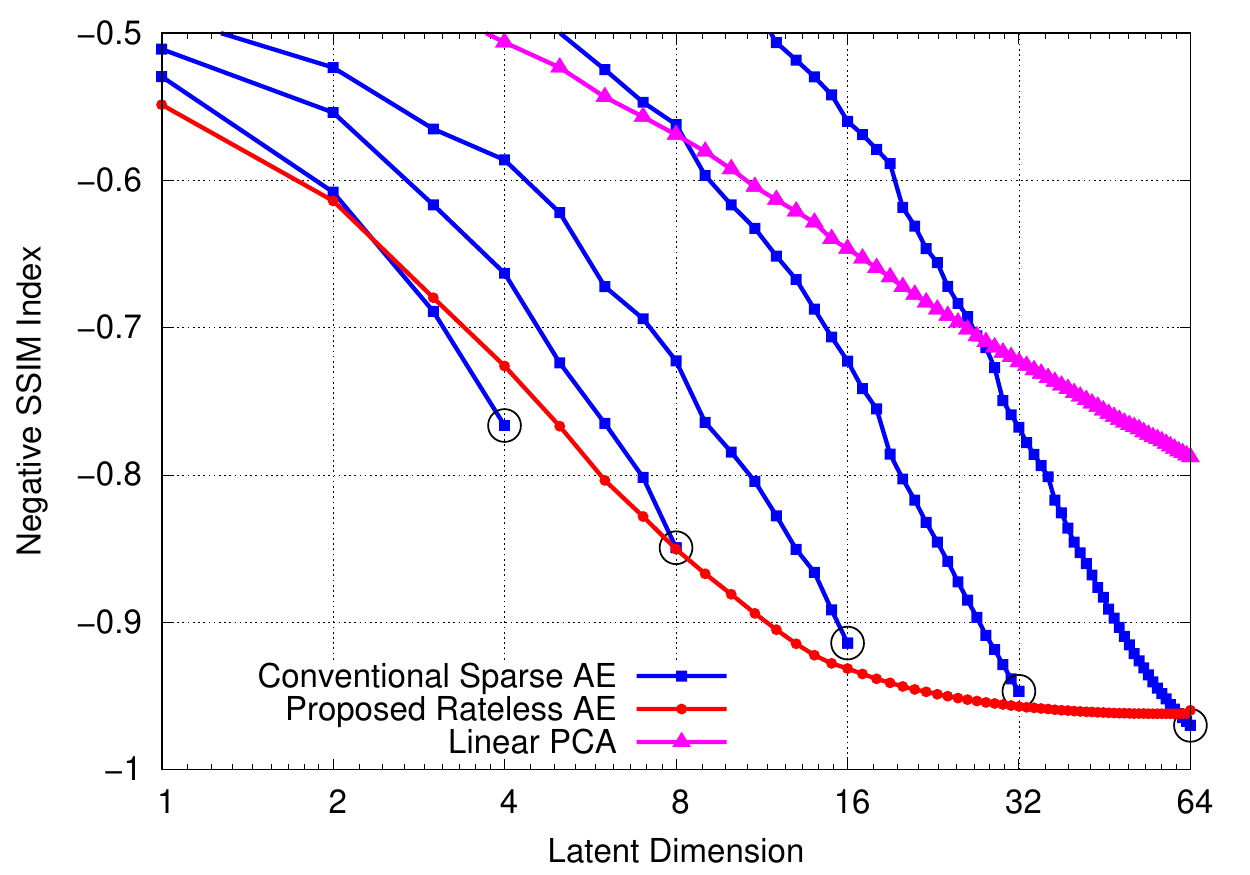}}
 \subfloat[CIFAR-10]{\includegraphics[width=0.49\linewidth]{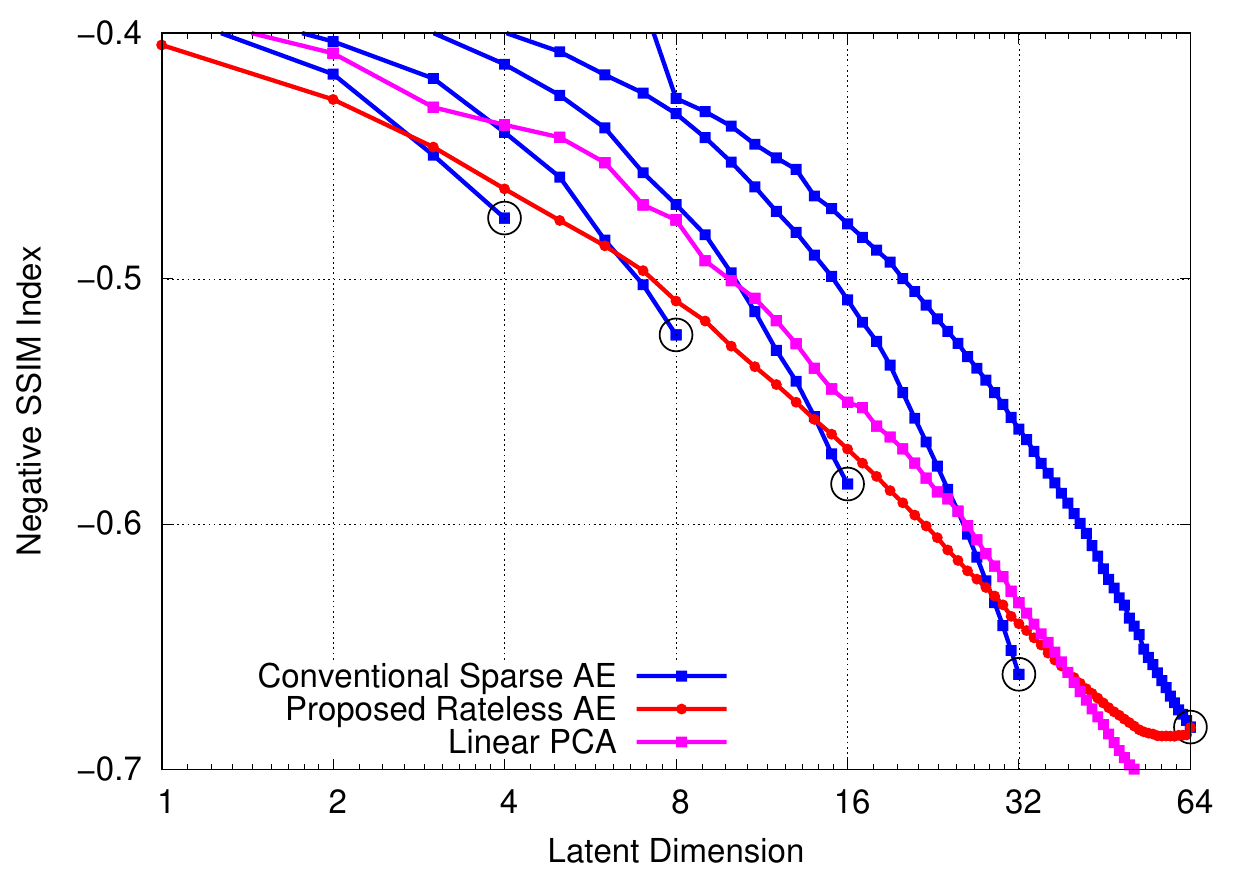}}
 \caption{SSIM performance of SAE and RL-AE as a function of survivor latent dimensionality $L$.}
 \label{fig:ssim}
\end{figure}

Figs.~\ref{fig:ssim}(a) and (b) plot the negative SSIM index of the reconstructed images by the conventional SAE and proposed RL-AE for MNIST and CIFAR-10 datasets, respectively.
It is confirmed in those figures that the conventional SAE cannot be universally used for flexibly varying dimensionality in the SSIM distortion metric.
Although the proposed RL-AE may perform worse than the conventional SAEs at some dimensionalities, for which the SAE models were dedicatedly optimized,
our RL-AE flexibly achieves SSIM performance closely comparable to the best SSIMs obtained by the ensemble of SAEs over the wide range of dimensionalities $L \leq 64$.

We can also see that the traditional PCA has a higher loss in the perceptual SSIM metric compared to the MSE metric.
In particular for MNIST in Fig.~\ref{fig:ssim}(a), the SSIM degradation of the PCA over our RL-AE is noticeable over the whole range of dimensionalities, while the PCA worked well for lower dimensionality for the MSE metric, as seen in Fig.~\ref{fig:mse}(a).
More importantly, our AE can offer a perceptual performance benefit in the SSIM metric over PCA even for CIFAR-10 datasets, for which the AEs could not outperform the PCA in the MSE metric as discussed in Fig.~\ref{fig:mse}(b).
This makes sense because the PCA does not consider any perceptual quality but only the signal energy relevant for the MSE measure.

\subsection{Reconstructed images}

\begin{figure}[t]
 \centering
 \subfloat[Conventional Sparse AE]{\includegraphics[width=0.47\linewidth,trim=1.6in 1.3in 1.6in 0.7in,clip]{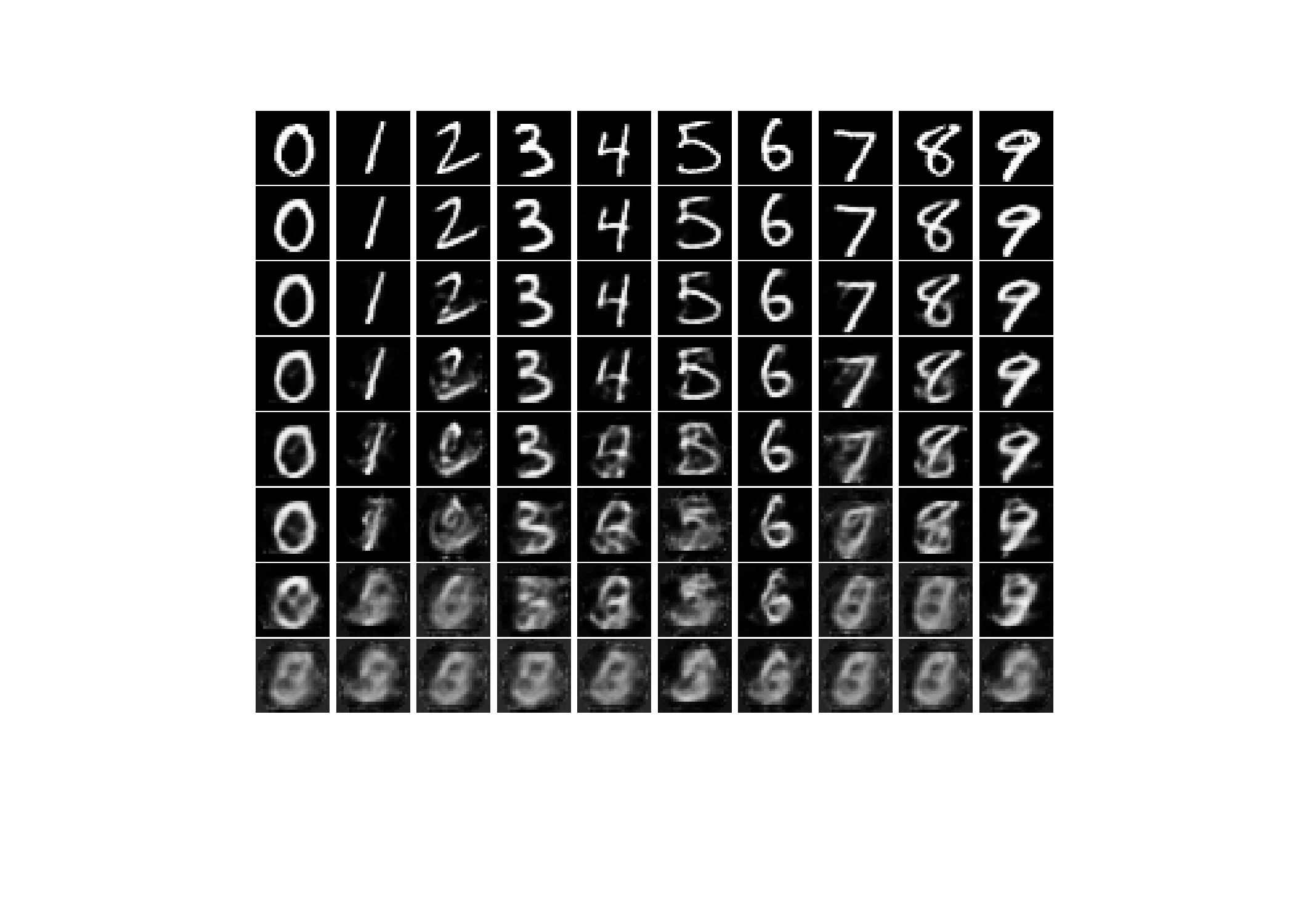}}
 \hfil
 \subfloat[Proposed Rateless AE]{\includegraphics[width=0.47\linewidth,trim=1.6in 1.3in 1.6in 0.7in,clip]{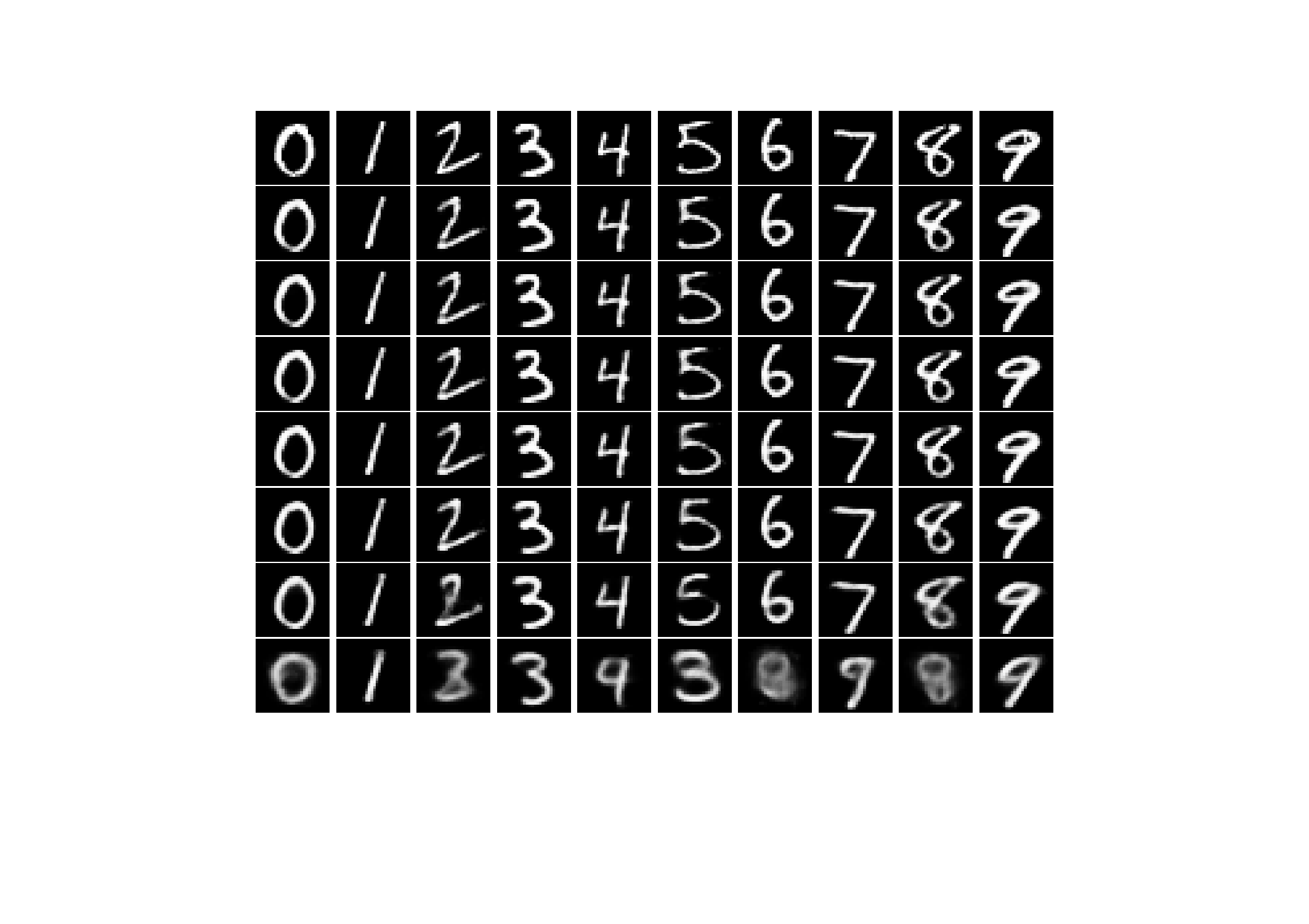}}
 \caption{MNIST reconstruction snapshots varying the survivor latent dimensionality $L$ using AE model designed at dimensionality of $M=64$. The top row is the original image, and subsequent rows are reconstructed images for a reduced dimensionality of $L=\{64, 54, 44, 34, 24, 14, 4\}$. }
 \label{fig:mnist}
\end{figure}

\begin{figure}[t]
 \centering
 \subfloat[Conventional Sparse AE]{\includegraphics[width=0.47\linewidth,trim=1.6in 1.3in 1.6in 0.7in,clip]{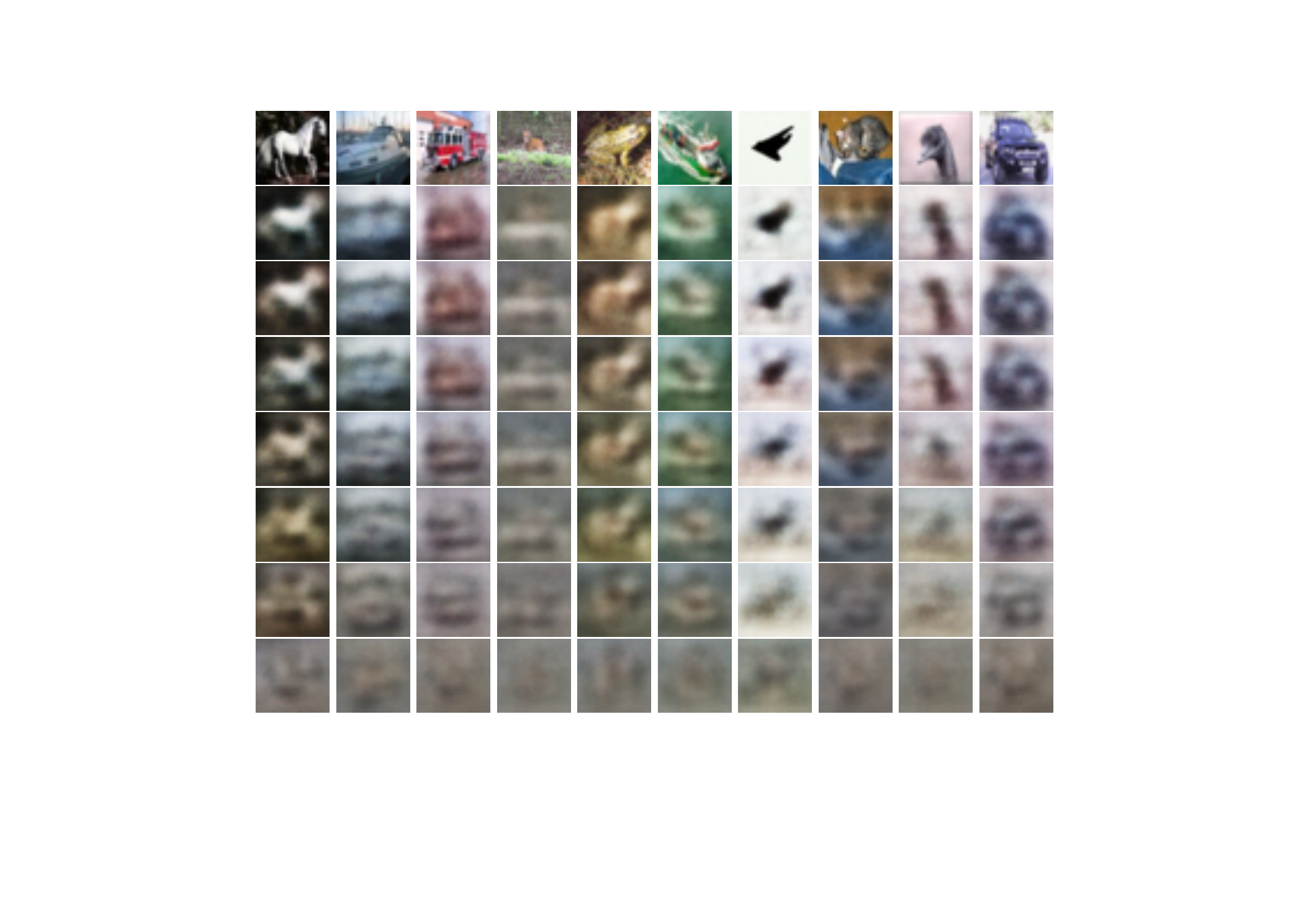}}
 \hfil
 \subfloat[Proposed Rateless AE]{\includegraphics[width=0.47\linewidth,trim=1.6in 1.3in 1.6in 0.7in,clip]{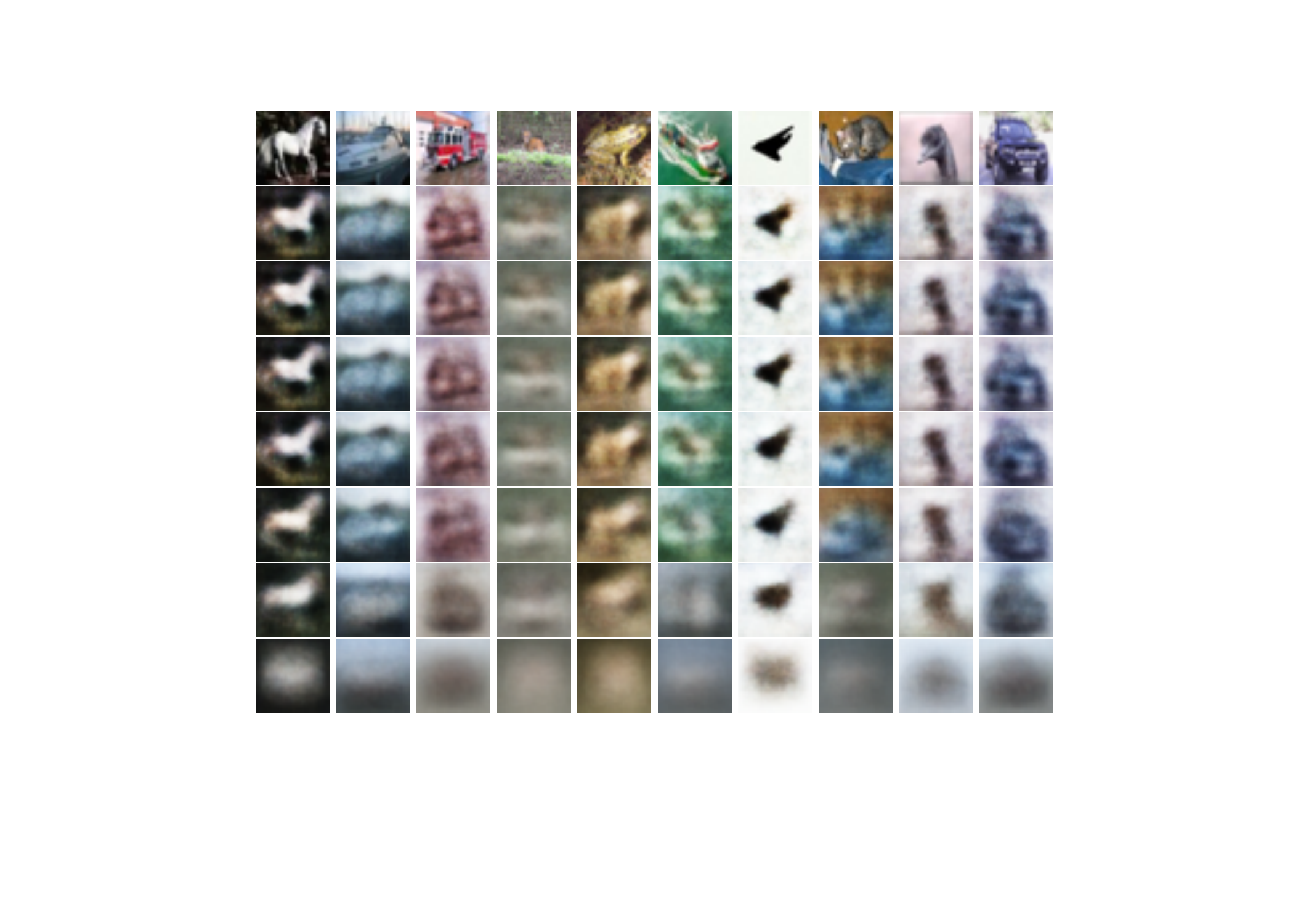}}
 \caption{CIFAR-10 reconstruction snapshots varying the survivor latent dimensionality $L$ using AE model designed at dimensionality of $M=64$. The top row is the original image, and subsequent rows are reconstructed images for a reduced dimensionality of $L=\{64, 54, 44, 34, 24, 14, 4\}$. }
 \label{fig:cifar10}
\end{figure}

Figs.~\ref{fig:mnist}(a) and (b) show visual samples randomly chosen from MNIST test datasets, respectively for SAE and RL-AE reconstructions.
The top row displays the original MNIST images, and the subsequent rows are reconstructed images for a reduced dimensionality of $L = \{64, 54, 44, 34, 24, 14, 4\}$.
Both types of models are trained at a latent dimensionality of $M=64$ under the MSE measure.
Our proposed RL-AE clearly exhibits improved visual quality for flexible dimensionality reduction versus the conventional SAE, without requiring retraining for each reduced dimensionality.
Similar results can be seen for CIFAR-10 in Figs.~\ref{fig:cifar10}(a) and (b).

Tables~\ref{mnist} and~\ref{cifar10} show the corresponding averaged MSE and SSIM index performance at $L=\{64, 54, \ldots, 4\}$ for MNIST and CIFAR-10, respectively.
Here, we also present the $10$-label classification accuracy when a classical support vector machine (SVM) is applied to the reduced-dimension latent variables.
Besides the higher image quality, we also observe higher classification accuracy achieved by the proposed rateless AE across the variable dimensionality.

\begin{table}
  \caption{MSE, SSIM, and SVM classification accuracy of SAE and RL-AE, optimized under MSE measure at dimensionality of $M=64$ for MNIST datasets}
  \label{mnist}
  \centering
  \begin{tabular}{ll rrrrrrr}
   \toprule
   \multicolumn{2}{l}{Dimensionality $L$}  & $64$ & $54$ & $44$ & $34$ & $24$ & $14$ & $4$ \\
   \midrule
   \multirow{2}{*}{MSE (dB)} & Conv. AE & $\mathbf{-5.81}$ & $-2.33$ & $0.56$ & $3.01$ & $5.03$ & $6.91$ & $8.18$ \\
                             & Prop. AE & $-5.19$ & $\mathbf{-5.26}$ & $\mathbf{-5.00}$ & $\mathbf{-4.35}$ & $\mathbf{-3.00}$ & $\mathbf{-0.05}$ & $\mathbf{5.16}$\\
   \midrule
   \multirow{2}{*}{SSIM Index} & Conv. AE & $\mathbf{0.97}$ & $0.91$ & $0.80$ & $0.66$ & $0.48$ & $0.25$ & $0.11$ \\
                             & Prop. AE & $\mathbf{0.97}$ & $\mathbf{0.97}$ & $\mathbf{0.97}$ & $\mathbf{0.96}$ & $\mathbf{0.95}$ & $\mathbf{0.89}$ & $\mathbf{0.66}$\\
   \midrule
   \multirow{2}{*}{SVM Acc.} & Conv. AE & $\mathbf{0.99}$ & $0.98$ & $0.97$ & $0.96$ & $0.92$ & $0.79$ & $0.42$ \\
                             & Prop. AE & $\mathbf{0.99}$ & $\mathbf{0.99}$ & $\mathbf{0.99}$ & $\mathbf{0.99}$ & $\mathbf{0.99}$ & $\mathbf{0.98}$ & $\mathbf{0.73}$\\
   \bottomrule
  \end{tabular}
\end{table}

\begin{table}
  \caption{MSE, SSIM, and SVM classification accuracy of SAE and RL-AE, optimized under MSE measure at dimensionality of $M=64$ for CIFAR-10 datasets}
  \label{cifar10}
  \centering
  \begin{tabular}{ll rrrrrrr}
   \toprule
   \multicolumn{2}{l}{Dimensionality $L$}  & $64$ & $54$ & $44$ & $34$ & $24$ & $14$ & $4$ \\
   \midrule
   \multirow{2}{*}{MSE (dB)} & Conv. AE & $-5.92$ & $-4.96$ & $-3.96$ & $-2.97$ & $-1.91$ & $-0.96$ & $0.92$ \\
                             & Prop. AE & $\mathbf{-6.19}$ & $\mathbf{-6.43}$ & $\mathbf{-6.30}$ & $\mathbf{-5.82}$ & $\mathbf{-5.11}$ & $\mathbf{-4.05}$ & $\mathbf{-1.92}$\\
   \midrule
   \multirow{2}{*}{SSIM Index} & Conv. AE & $0.64$ & $0.61$ & $0.57$ & $0.53$ & $0.48$ & $0.44$ & $0.37$ \\
                             & Prop. AE & $\mathbf{0.66}$ & $\mathbf{0.67}$ & $\mathbf{0.67}$ & $\mathbf{0.64}$ & $\mathbf{0.60}$ & $\mathbf{0.54}$ & $\mathbf{0.44}$\\
   \midrule
   \multirow{2}{*}{SVM Acc.} & Conv. AE & $\mathbf{0.47}$ & $0.47$ & $0.46$ & $0.44$ & $0.40$ & $0.32$ & $0.20$ \\
                             & Prop. AE & $\mathbf{0.47}$ & $\mathbf{0.48}$ & $\mathbf{0.47}$ & $\mathbf{0.48}$ & $\mathbf{0.46}$ & $\mathbf{0.42}$ & $\mathbf{0.29}$\\
   \bottomrule
  \end{tabular}
\end{table}

\subsection{Latent representation}

\begin{figure}[t]
 \centering
 \subfloat[Conventional Sparse AE]{\includegraphics[width=0.49\linewidth]{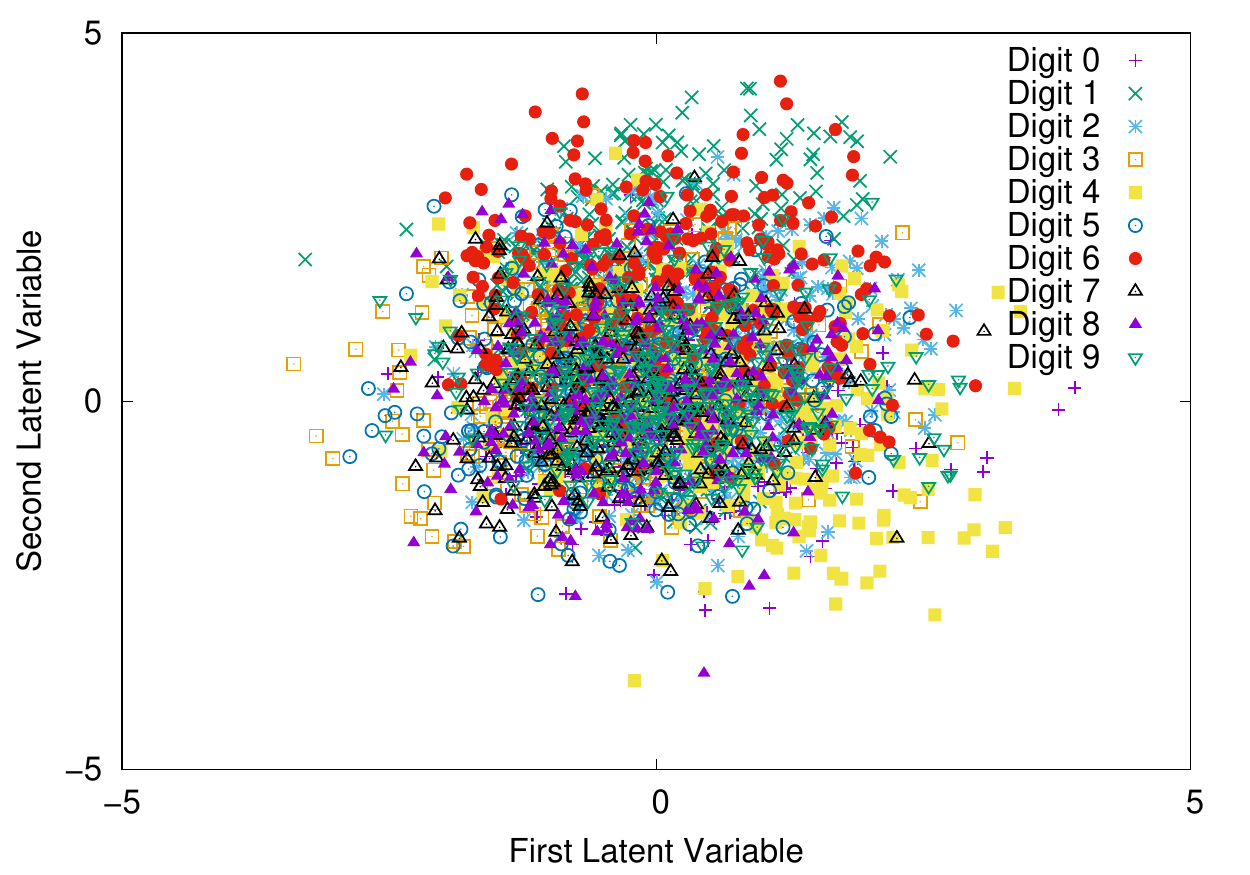}}
 \subfloat[Proposed Rateless AE]{\includegraphics[width=0.49\linewidth]{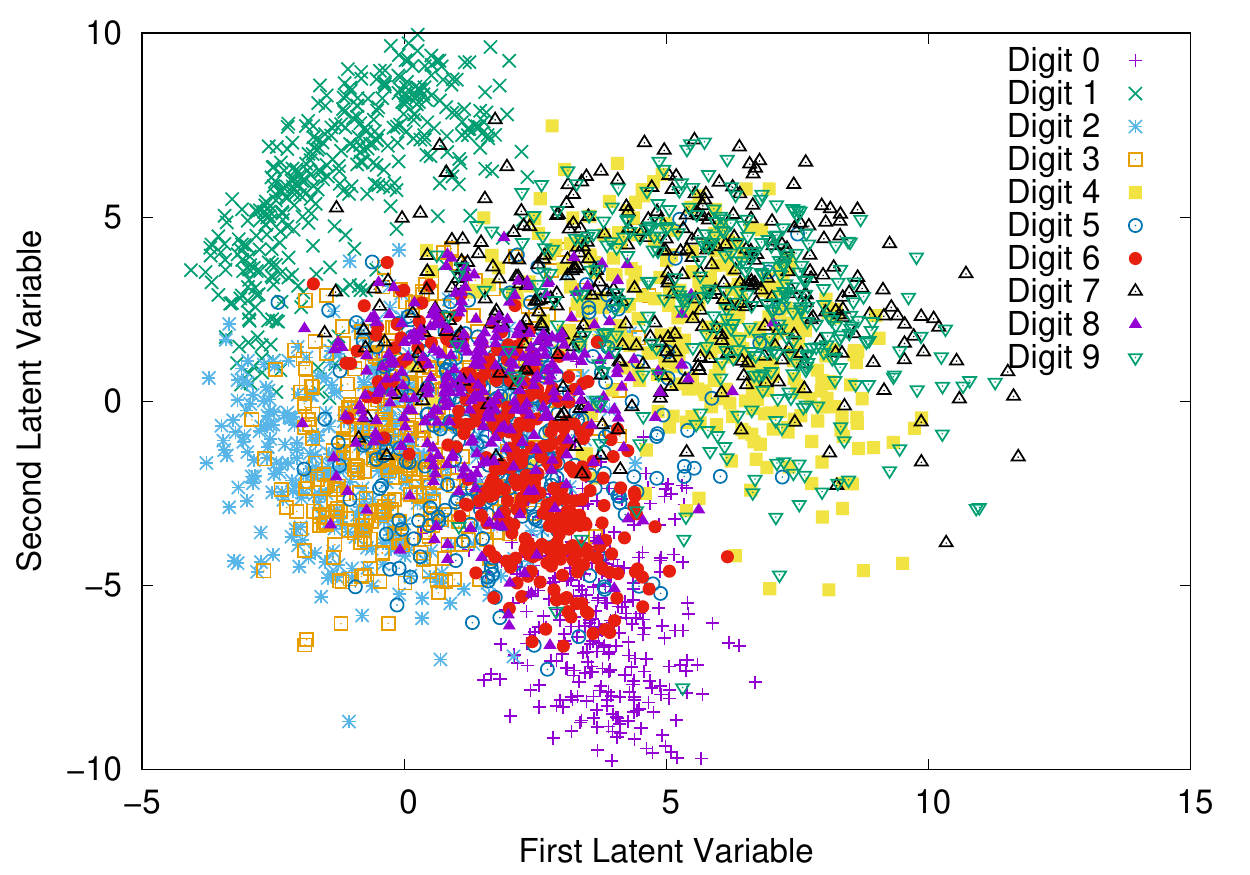}}
 \caption{The first two latent variables of MNIST test images encoded by SAE and RL-AE, which are trained at a latent dimensionality of $M=64$ in terms of MSE measure.}
 \label{fig:latent}
\end{figure}

Finally we show a latent space geometry in Figs.~\ref{fig:latent}(a) and (b), where the first two latent variables of all MNIST test images are plotted for the traditional SAE and proposed RL-AE, respectively.
One can clearly see that the label-dependent distribution in our RL-AE is more clearly observable than the conventional AE, since the most-principal latent components are properly associated with the upper latent variables via the proposed stochastic bottleneck technique.
This observation is expected from the higher SVM accuracy performance in Table~\ref{mnist}.

\section{Conclusions}

We proposed new a type of auto-encoders employing a form of stochastic bottlenecking with non-uniform dropout rates for flexible dimensionality reduction.
The proposed auto-encoders are rateless, i.e., the compression rate in dimensionality reduction is not pre-determined at the training phase and the user can freely change the dimensionality at testing phase without severely degrading quality.
To realize rateless AEs, a simple regularization method called TailDrop was introduced to impose higher priority at upper neurons for learning the most-principal nonlinear features.
This paper showed proof-of-concept results based on the standard MNIST and CIFAR-10 image datasets.
Universally good distortion performance was obtained with a single AE model irrespective of the flexible dimensionality reduction rate, which was obtained by simply dropping the least-principal latent dimensions.
More rigorous analysis and theoretical optimization of dropout rate distributions for real-world data are left for future work.
Multi-objective learning to account for various downstream applications is also an important open question to pursue.

%



\section{Supplementary Experiments}

We show the MSE performance of the proposed RL-AE for different datasets as follows:
\begin{itemize}
 \item Fashion-MNIST (FMNIST)~\cite{Xiao:2017} is a set of fashion articles represented by gray-scale $28$-by-$28$ images, associated with a label from $10$ classes, consisting a training set of $60{,}000$ examples and a test set of $10{,}000$ examples.
FMNIST was intended to serve as a direct replacement for the MNIST dataset for benchmarking.
 \item Kuzushiji-MNIST (KMNIST)~\cite{Clanuwat:2018} is another set of hand-written Japanese characters represented by $10$-class gray-scale $28$-by-$28$ images with the same data sizes of MNIST and FMNIST.
 \item The street view house numbers (SVHN) dataset~\cite{Netzer:2011} is similar to MNIST but composed of cropped $32$-by-$32$ color images of house numbers. It contains $73{,}257$ digits for training and $26{,}032$ digits for testing.
 \item CIFAR-100~\cite{Krizhevsky:2009} is a set of small natural images, just like the CIFAR-10, except it has $100$ classes containing $600$ images each. There are $500$ training images and $100$ testing images per class. The $100$ classes in the CIFAR-100 are grouped into $20$ superclasses.
\end{itemize}

\begin{figure}[t]
 \centering
 \subfloat[FMNIST]{\includegraphics[width=0.50\linewidth]{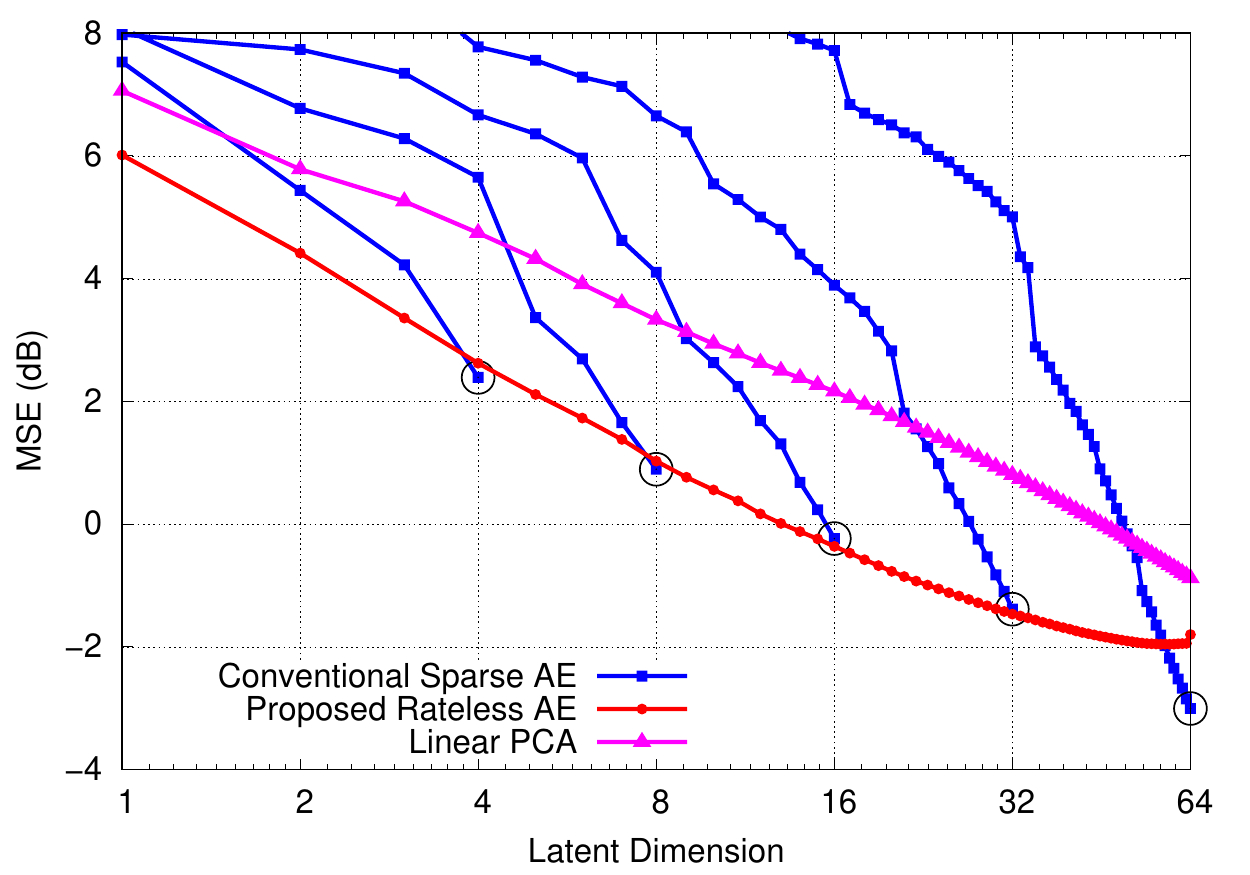}}
 \subfloat[KMNIST]{\includegraphics[width=0.50\linewidth]{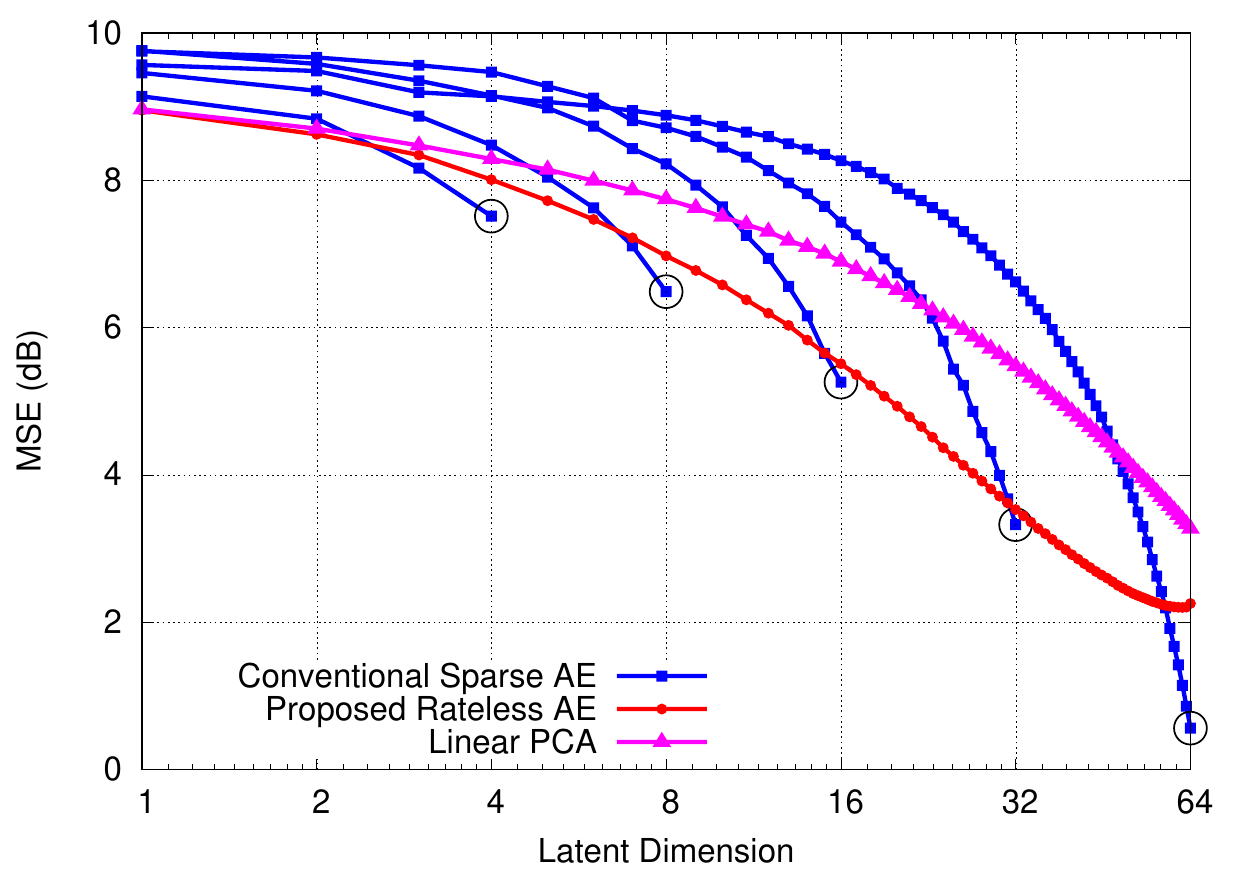}}\\
 \subfloat[SVHN]{\includegraphics[width=0.50\linewidth]{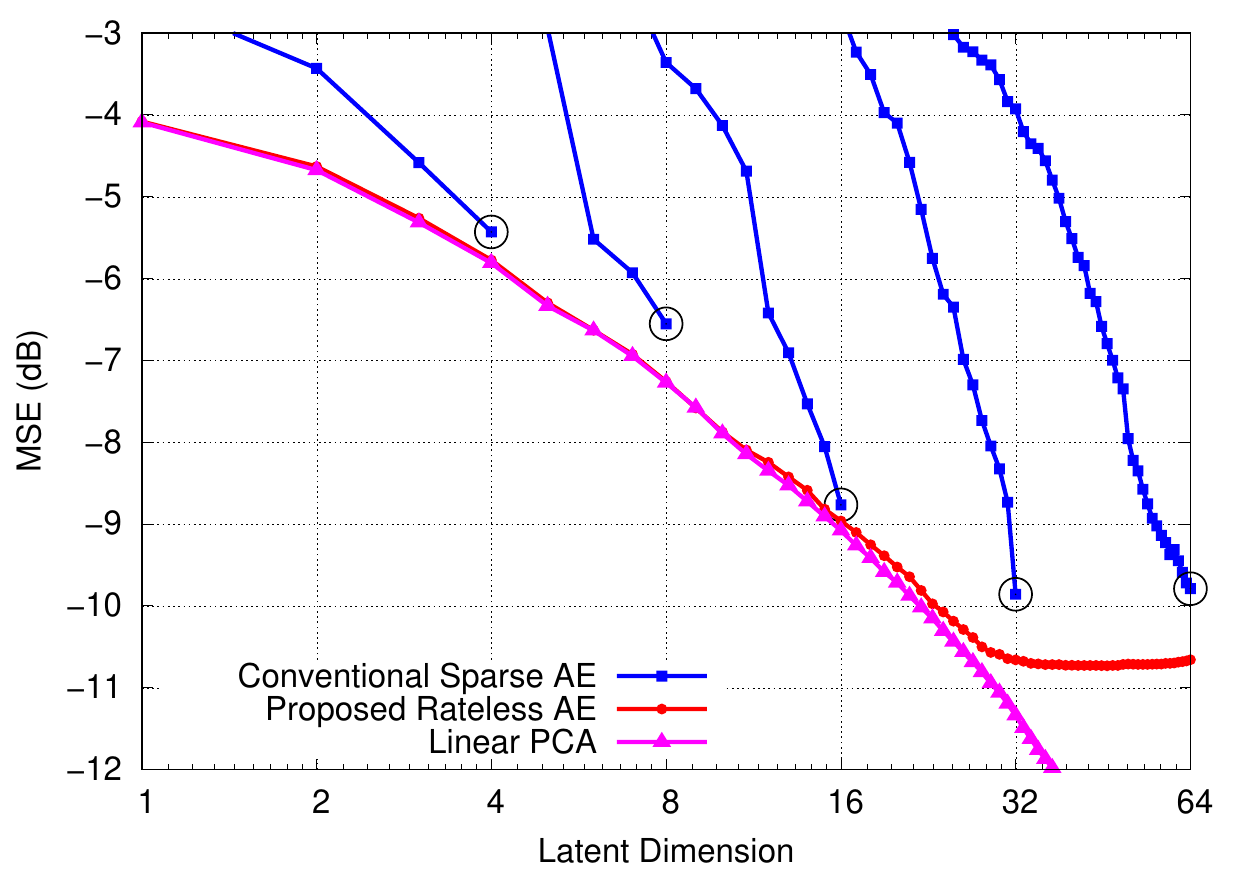}}
 \subfloat[CIFAR-100]{\includegraphics[width=0.50\linewidth]{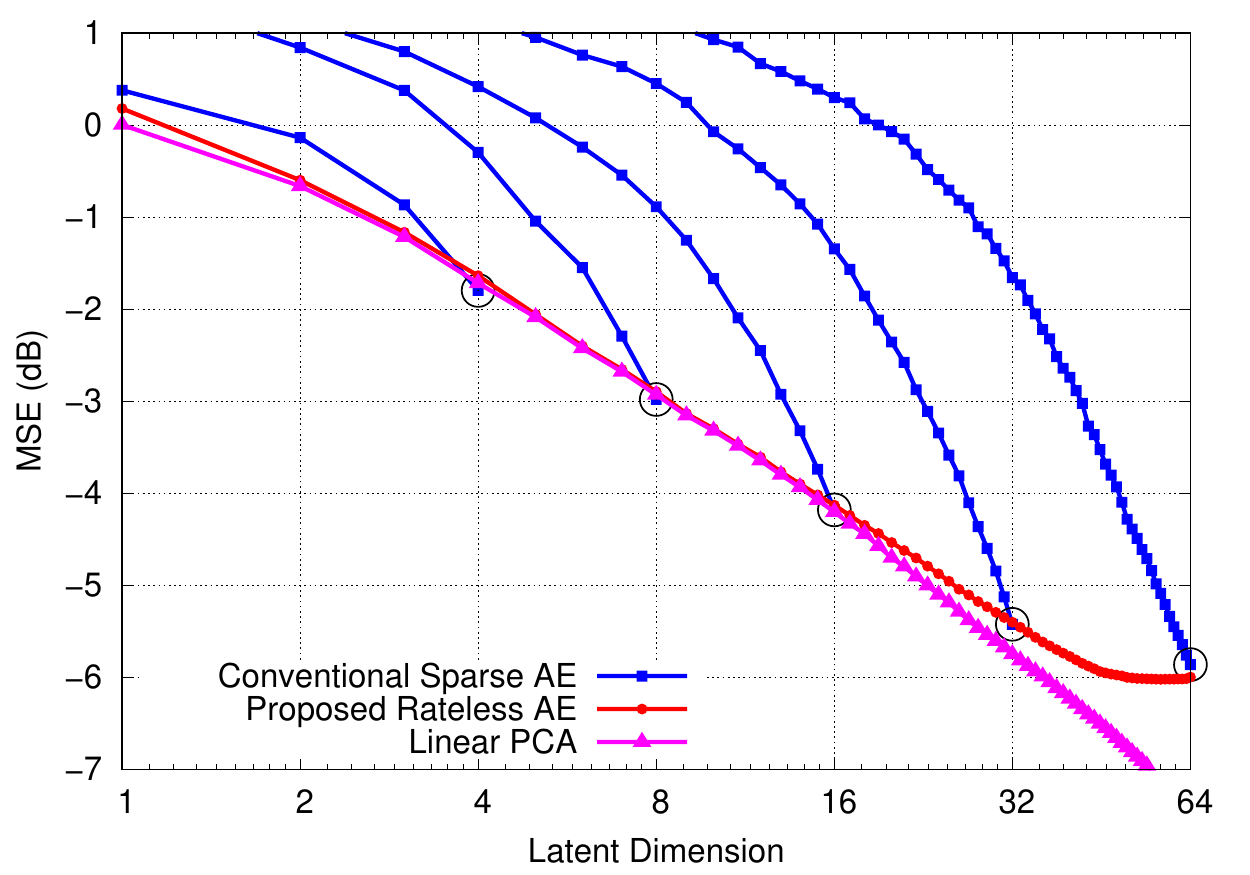}}
 \caption{MSE performance of SAE and RL-AE as a function of survivor latent dimensionality $L$.}
 \label{fig:mse2}
\end{figure}

Figs.~\ref{fig:mse2}(a) through (d) show the MSE performance as a function of survivor latent dimensionality $L$ for FMNIST, KMNIST, SVHN, and CIFAR-100, respectively.
We can confirm that the proposed AE achieves graceful performance over the wide range of dimensionality, competitive to the best performance which the conventional AEs can offer at a pre-determined dimensionality.
Although the linear PCA also achieves rateless performance, a significant MSE loss is seen for gray-scale datasets of FMNIST and KMNIST, similar to MNIST in Fig.~\ref{fig:mse}(a).
However for color datasets, PCA performed well just like in CIFAR-10 in Fig.~\ref{fig:mse}(b).
Nonetheless, our AE achieves nearly best performance, outperforming the conventional AE.
In addition, our AE may achieve better perceptual quality and classification accuracy as discussed for CIFAR-10.
The experimental results verified that a simple mechanism with non-uniform dropout regularization can enable a reasonable rateless property.

\begin{figure}[t]
 \centering
 \subfloat[Conventional Sparse AE]{\includegraphics[width=0.47\linewidth,trim=1.6in 1.3in 1.6in 0.7in,clip]{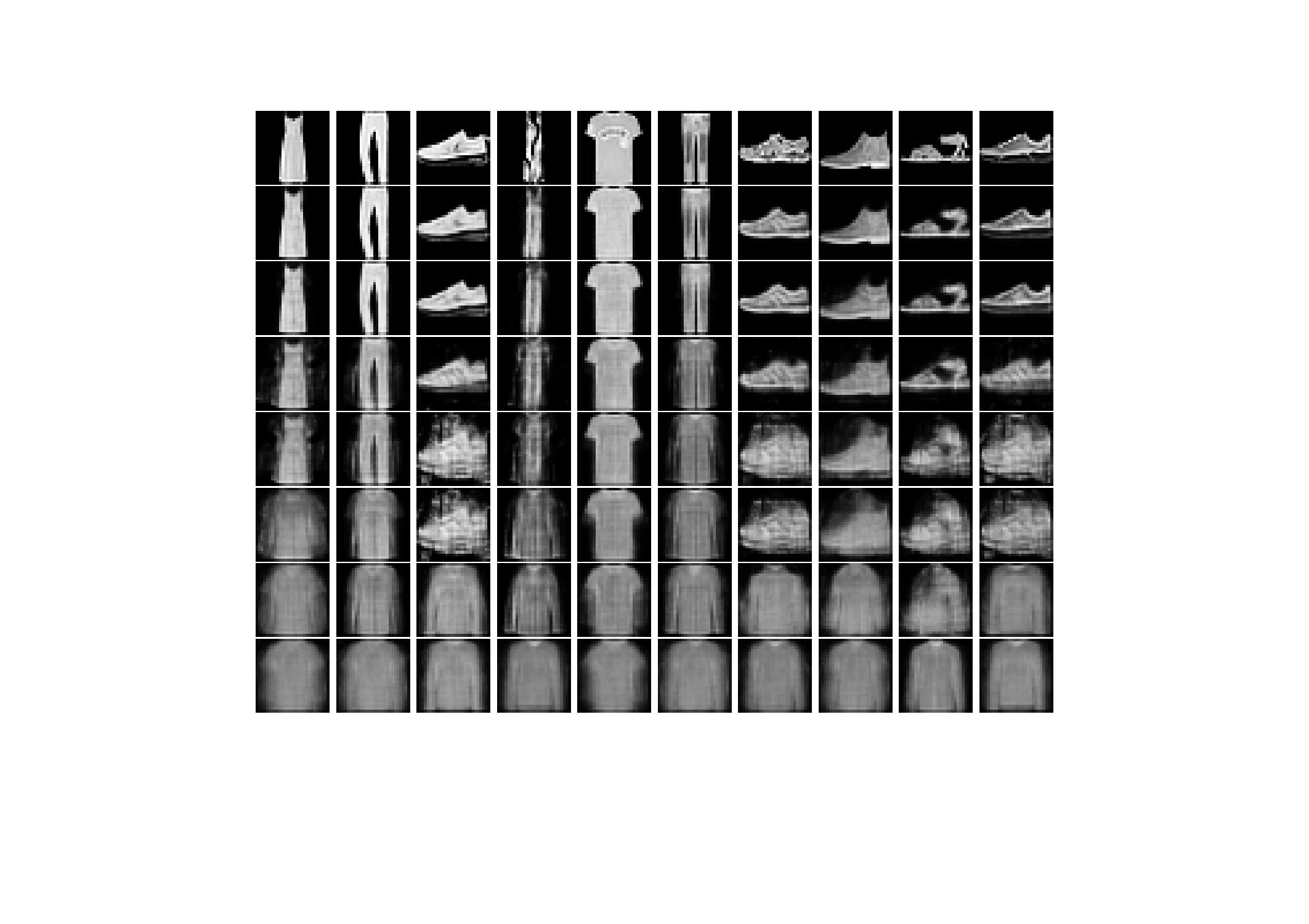}}
 \hfil
 \subfloat[Proposed Rateless AE]{\includegraphics[width=0.47\linewidth,trim=1.6in 1.3in 1.6in 0.7in,clip]{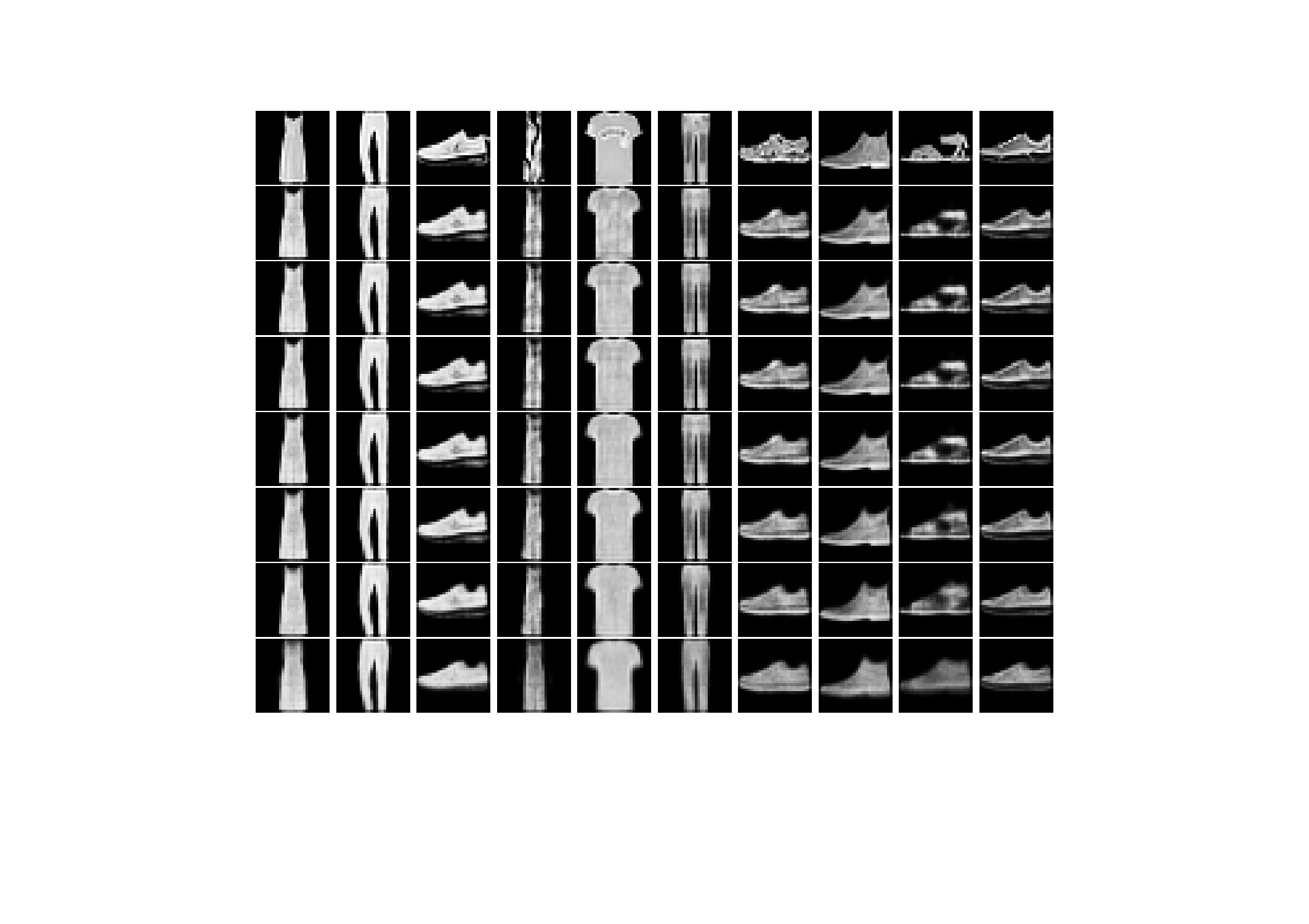}}
 \caption{FMNIST reconstruction snapshots varying the survivor latent dimensionality $L$ using AE model designed at dimensionality of $M=64$. The top row is the original image, and subsequent rows are reconstructed images for a reduced dimensionality of $L=\{64, 54, 44, 34, 24, 14, 4\}$. }
 \label{fig:fmnist}
\end{figure}

\begin{figure}[t]
 \centering
 \subfloat[Conventional Sparse AE]{\includegraphics[width=0.47\linewidth,trim=1.6in 1.3in 1.6in 0.7in,clip]{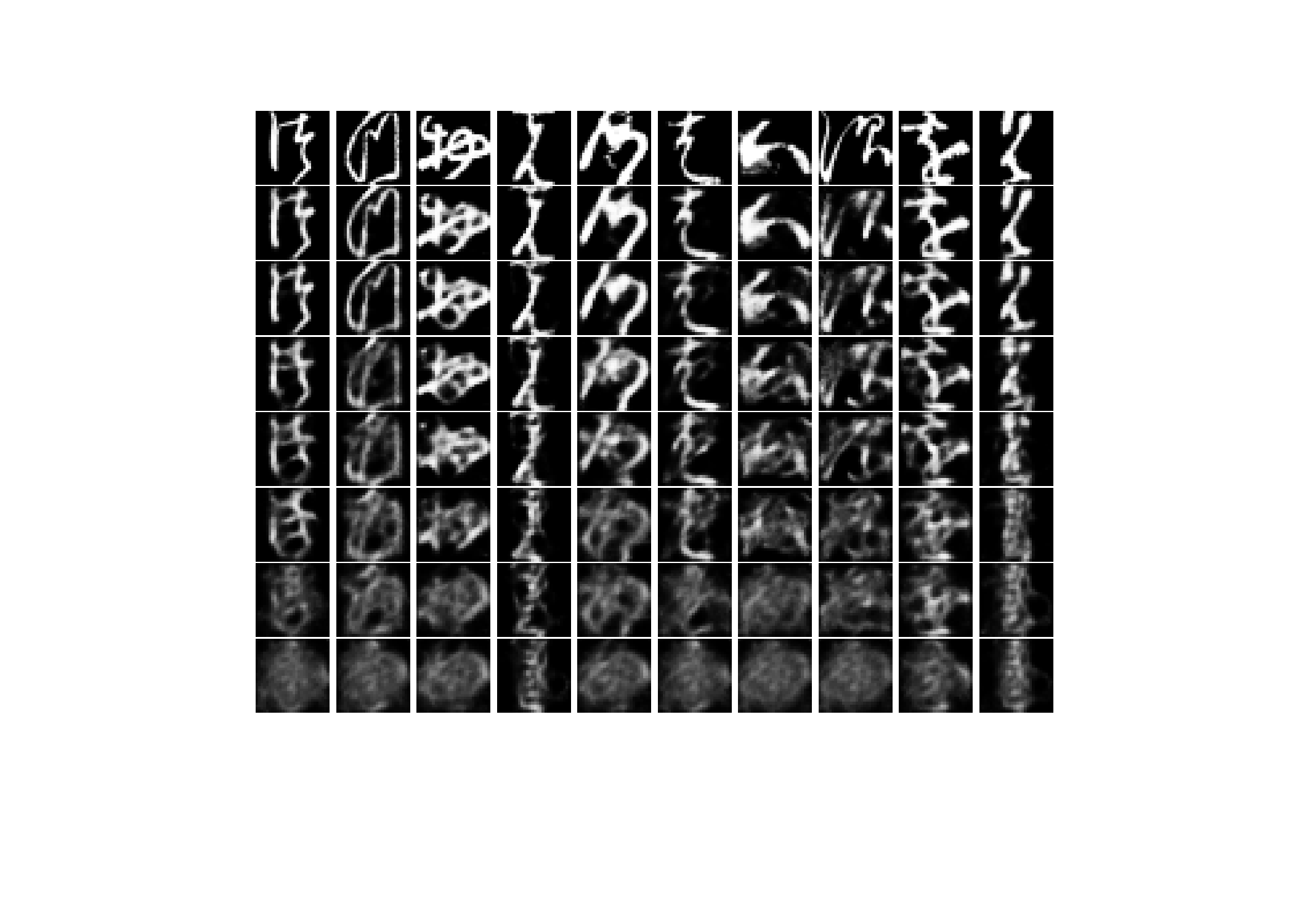}}
 \hfil
 \subfloat[Proposed Rateless AE]{\includegraphics[width=0.47\linewidth,trim=1.6in 1.3in 1.6in 0.7in,clip]{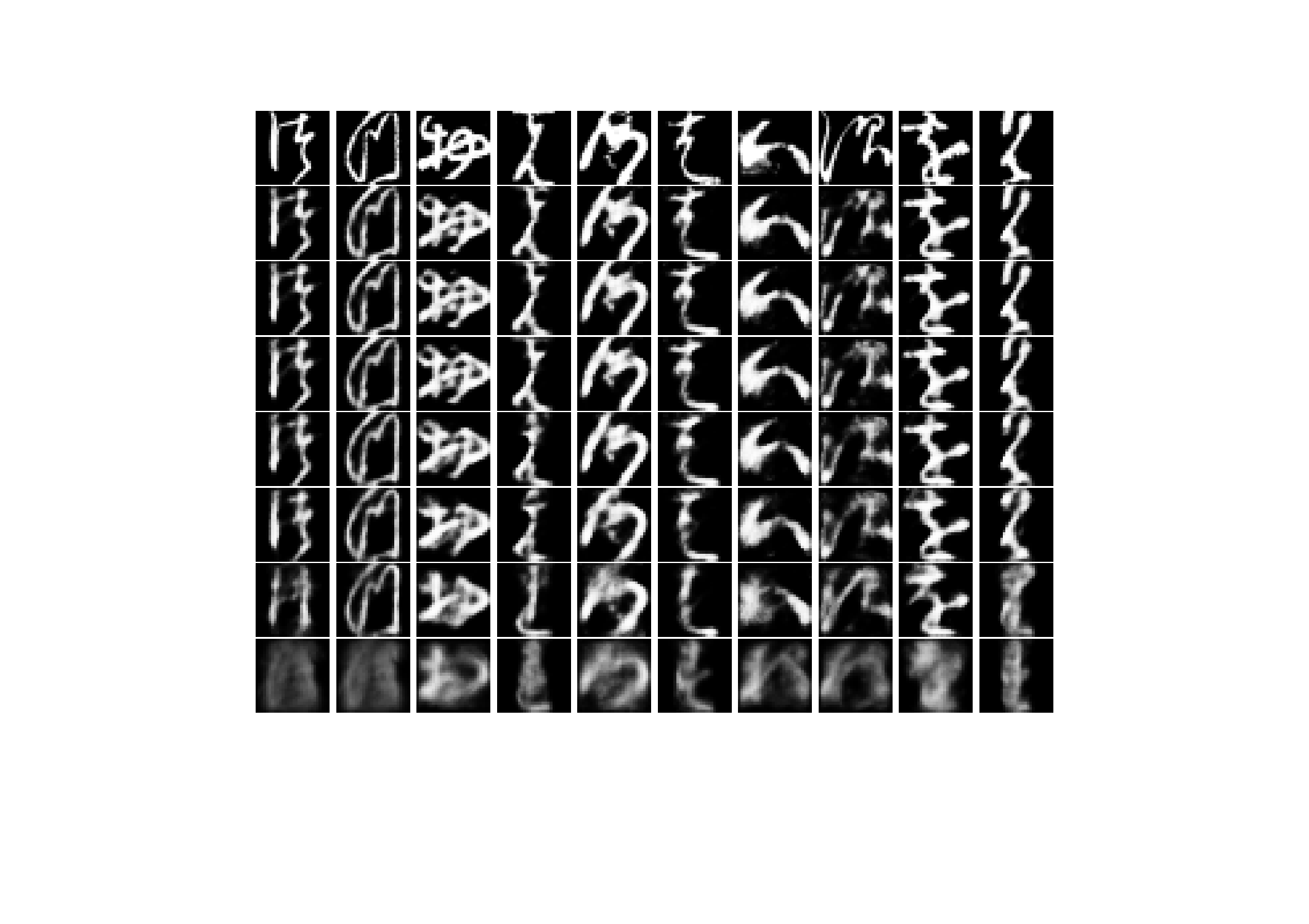}}
 \caption{KMNIST reconstruction snapshots varying the survivor latent dimensionality $L$ using AE model designed at dimensionality of $M=64$. The top row is the original image, and subsequent rows are reconstructed images for a reduced dimensionality of $L=\{64, 54, 44, 34, 24, 14, 4\}$. }
 \label{fig:kmnist}
\end{figure}

\begin{figure}[t]
 \centering
 \subfloat[Conventional Sparse AE]{\includegraphics[width=0.47\linewidth,trim=1.6in 1.3in 1.6in 0.7in,clip]{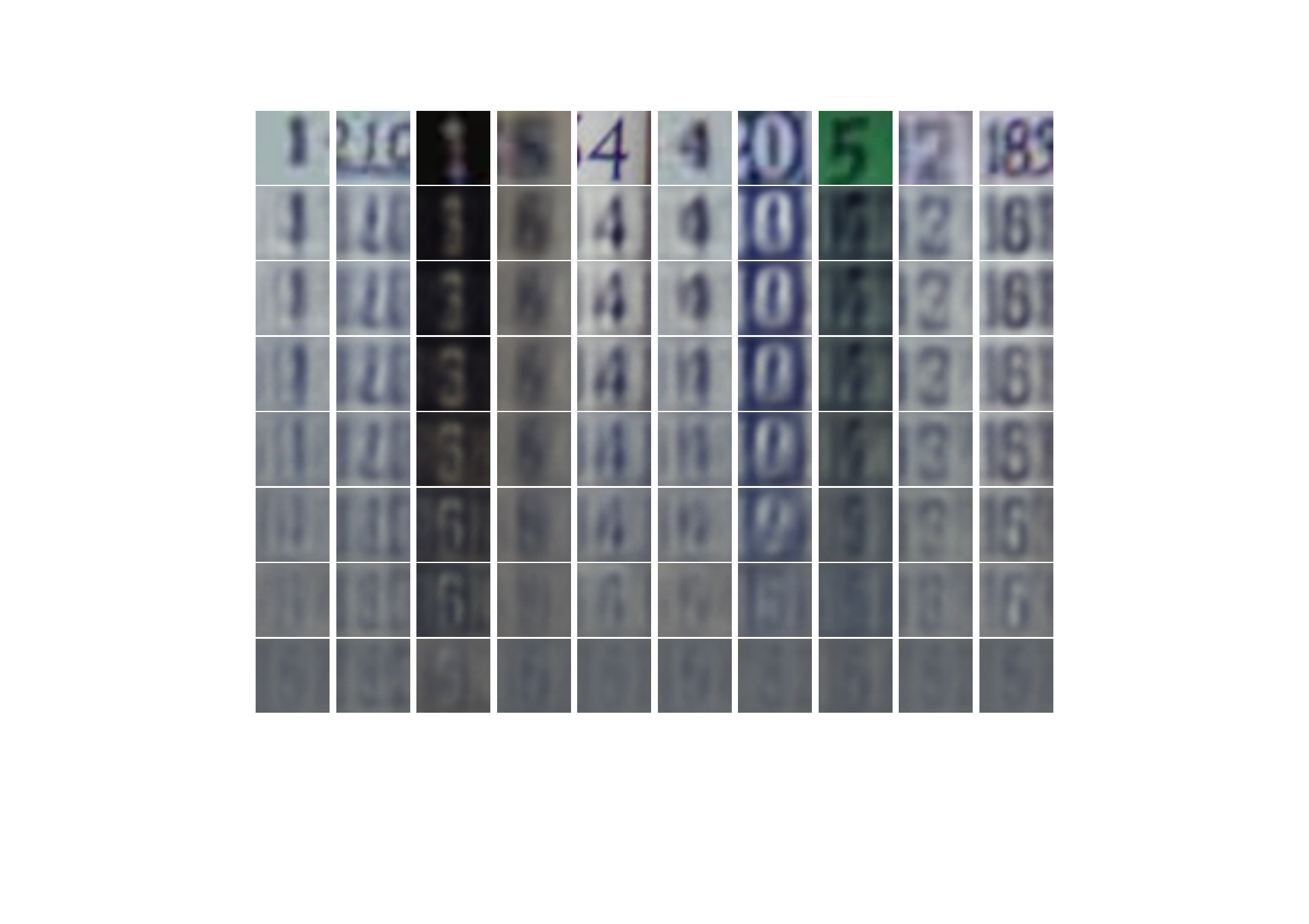}}
 \hfil
 \subfloat[Proposed Rateless AE]{\includegraphics[width=0.47\linewidth,trim=1.6in 1.3in 1.6in 0.7in,clip]{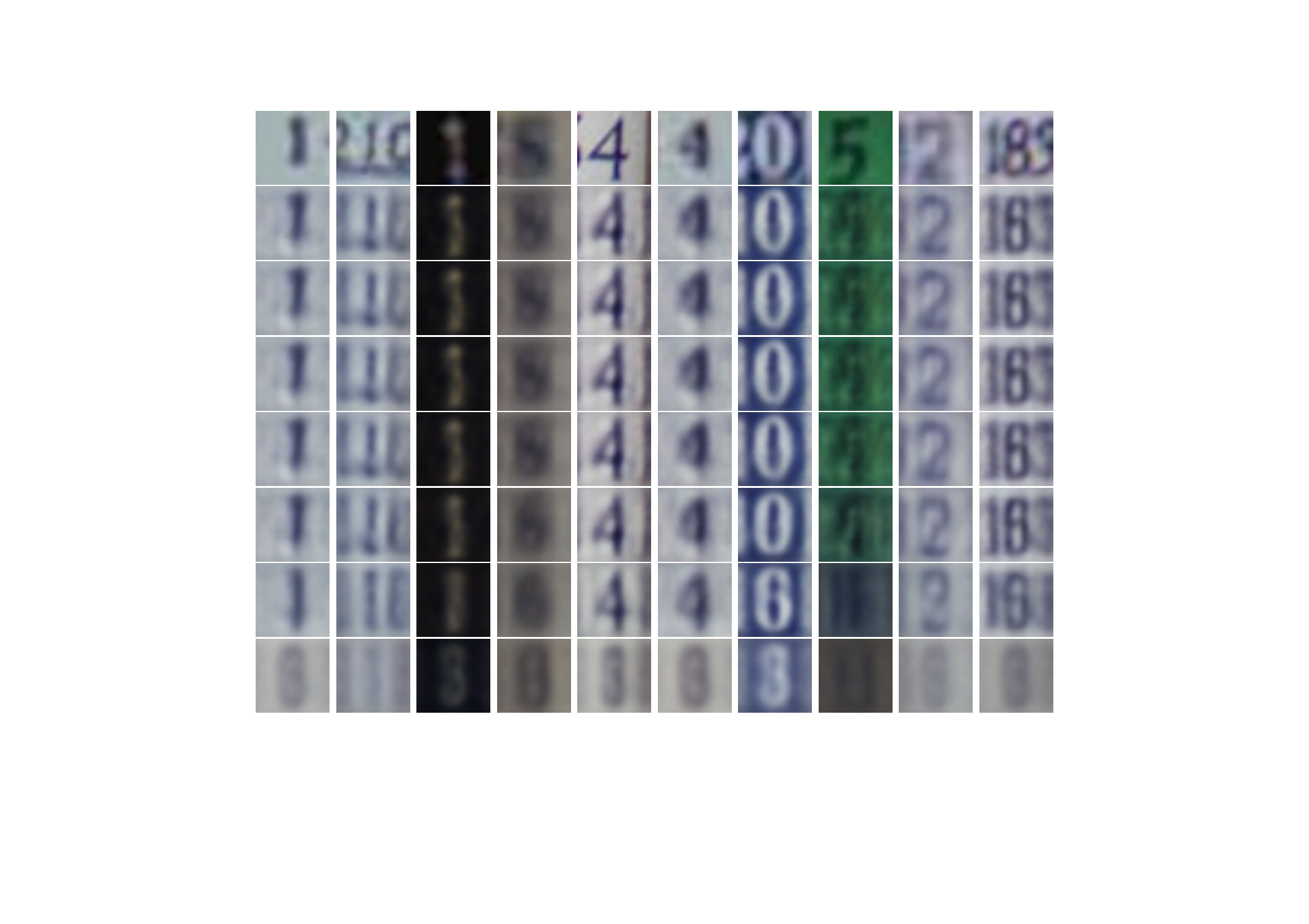}}
 \caption{SVHN reconstruction snapshots varying the survivor latent dimensionality $L$ using AE model designed at dimensionality of $M=64$. The top row is the original image, and subsequent rows are reconstructed images for a reduced dimensionality of $L=\{64, 54, 44, 34, 24, 14, 4\}$. }
 \label{fig:svhn}
\end{figure}

\begin{figure}[t]
 \centering
 \subfloat[Conventional Sparse AE]{\includegraphics[width=0.47\linewidth,trim=1.6in 1.3in 1.6in 0.7in,clip]{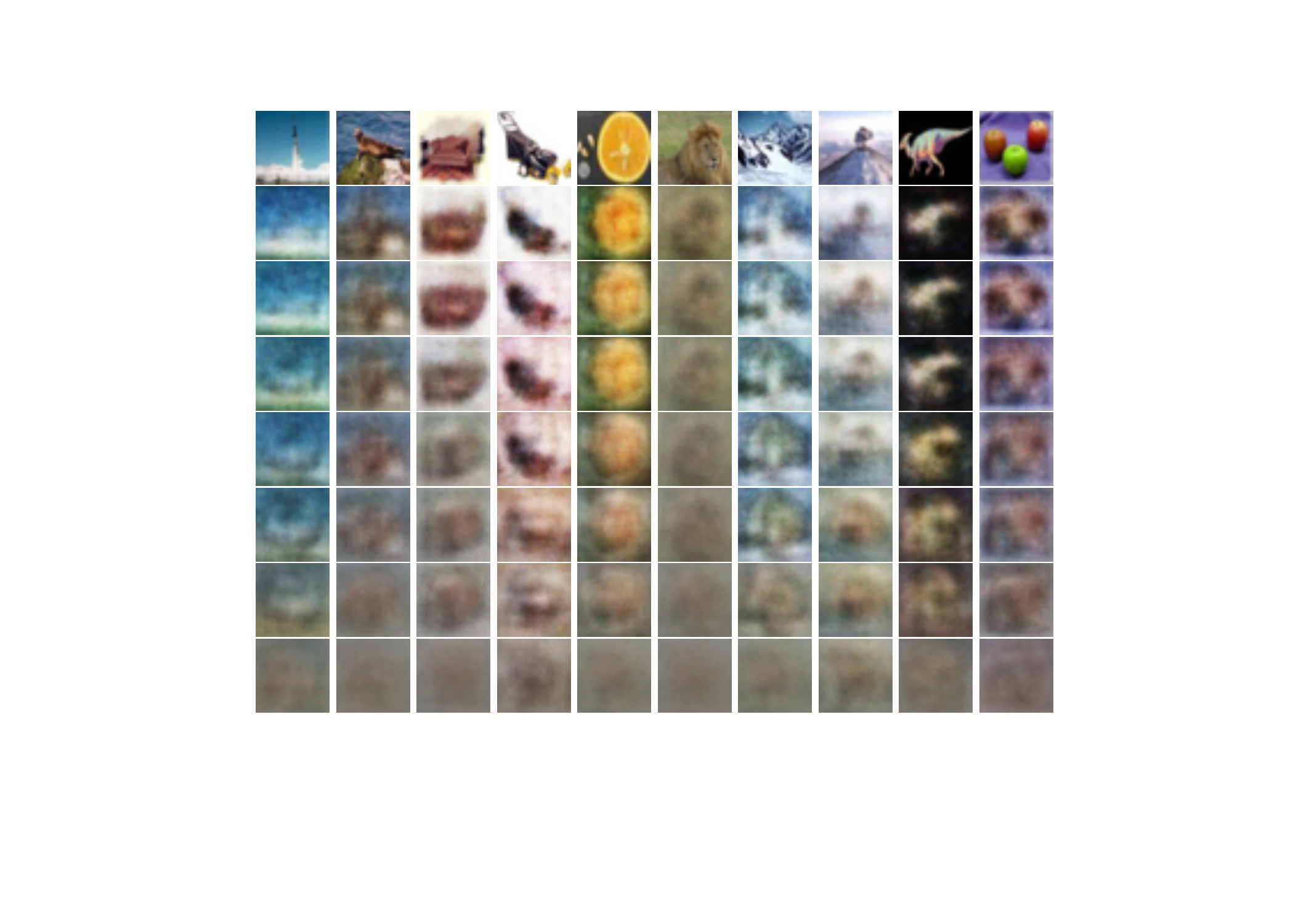}}
 \hfil
 \subfloat[Proposed Rateless AE]{\includegraphics[width=0.47\linewidth,trim=1.6in 1.3in 1.6in 0.7in,clip]{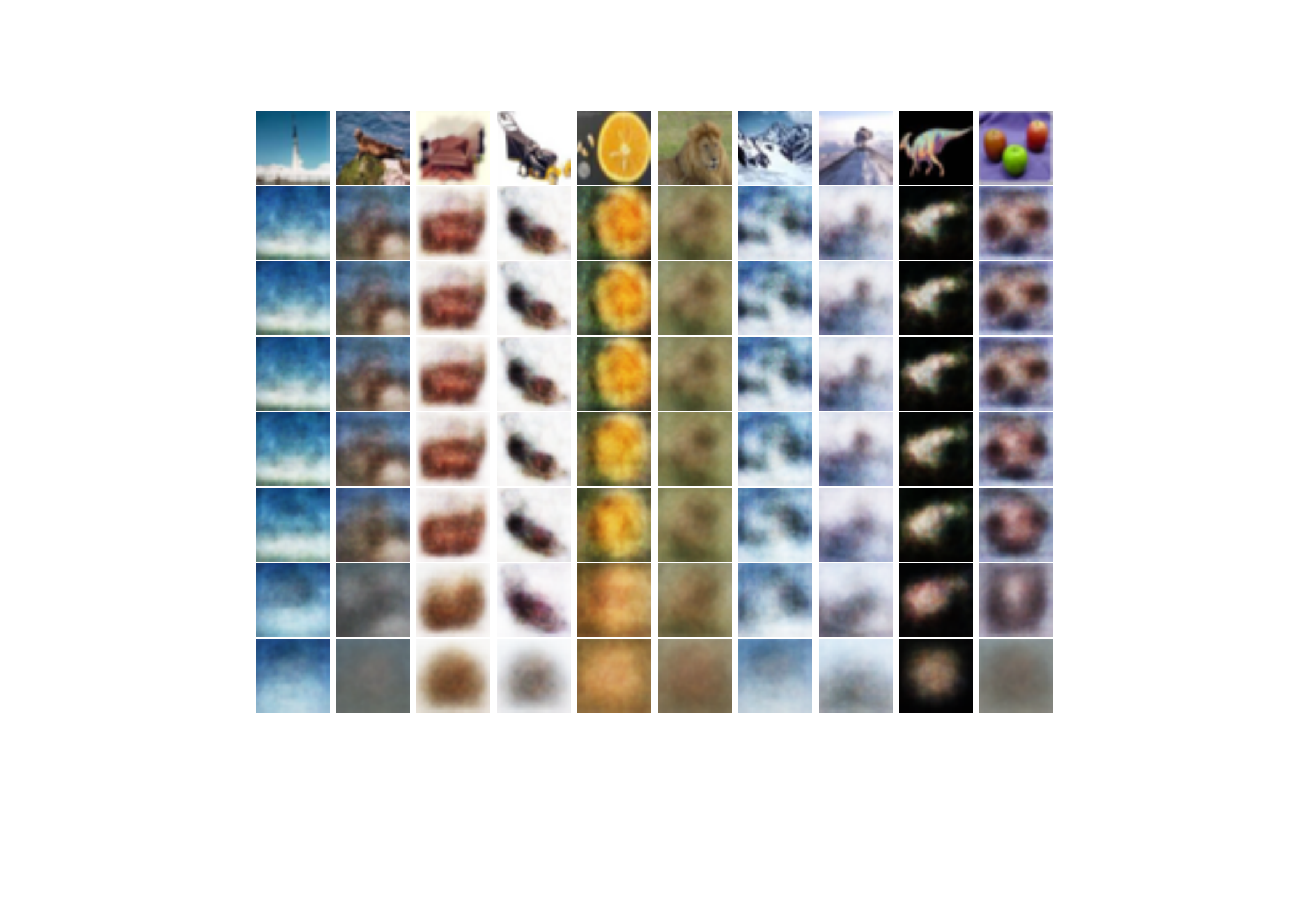}}
 \caption{CIFAR-100 reconstruction snapshots varying the survivor latent dimensionality $L$ using AE model designed at dimensionality of $M=64$. The top row is the original image, and subsequent rows are reconstructed images for a reduced dimensionality of $L=\{64, 54, 44, 34, 24, 14, 4\}$. }
 \label{fig:cifar100}
\end{figure}

Figs.~\ref{fig:fmnist}, \ref{fig:kmnist}, \ref{fig:svhn}, and \ref{fig:cifar100} show visual snapshots of randomly-chosen images reconstructed by the conventional AE and proposed AE for FMNIST, KMNIST, SVHN, and CIFAR-100, respectively.
One can observe a clear advantage of the RL-AE over the SAE to maintain higher quality across variable dimensionality.

\end{document}